\documentclass{article}

\PassOptionsToPackage{numbers}{natbib}
 \usepackage[preprint]{neurips_2026}

\usepackage[utf8]{inputenc} 
\usepackage[T1]{fontenc}    
\usepackage{hyperref}       
\usepackage{url}            
\usepackage{booktabs}       
\usepackage{amsfonts}       
\usepackage{nicefrac}       
\usepackage{microtype}      
\usepackage{xcolor}         

\usepackage{graphicx}
\usepackage{subcaption}
\usepackage{multirow}
\usepackage{array}
\usepackage{amsmath}
\usepackage{amssymb}

\title{Fitting Is Not Enough:\\Smoothness in Extremely Quantized LLMs}

\author{Yuzhuang Xu$^{1}$ \quad Xu Han$^{2,*}$ \quad Yuxuan Li$^{2}$ \quad Pengzhan Li$^{1}$ \quad Wanxiang Che$^{1,}$\thanks{Corresponding authors} \\
\textsuperscript{\rm 1}Harbin Institute of Technology, Harbin, China \\
\textsuperscript{\rm 2}Tsinghua University, Beijing, China \\
\texttt{\{xyz, car\}@ir.hit.edu.cn, han-xu@mail.tsinghua.edu.cn} }

\begin{document}

\maketitle

\begin{abstract}
Large language models (LLMs) achieve strong performance but incur high deployment costs, motivating extremely low-bit but lossy quantization. 
Existing quantization algorithms mainly focus on improving the numerical accuracy of forward computation to eliminate performance degradation.
In this paper, we show that extremely quantized LLMs suffer from systematic smoothness degradation beyond numerical precision loss. Through a smoothness proxy, we observe that such degradation becomes increasingly severe as the quantization bit-width decreases. Furthermore, based on sequence neighborhood modeling, we find that quantized models exhibit a rapid reduction of effective token candidates within the prediction neighborhood, which directly leads to a sparser decoding tree and degraded generation quality. To validate it, we introduce a simple smoothness-preserving principle in both post-training quantization and quantization-aware training, and demonstrate that preserving smoothness brings additional gains beyond numerical accuracy. The core goal of this paper is to highlight smoothness preservation as an important design consideration for future extreme quantization methods. Code is available at \url{https://github.com/xuyuzhuang11/FINE}.
\end{abstract}

{
\centering
\textbf{\textit{``When you reach the edge, hidden forces take over.''}}\par
}

\section{Introduction}

Large language models (LLMs) face substantial deployment costs, posing a major bottleneck for real-world adoption. 
To unlock model capability under constrained budgets, model quantization is widely used, replacing high-bit value representations with low-bit counterparts~\citep{gptq2022,spinquant2025}. 
Model quantization has now been pushed to an extremely low bit-width, such as 1-bit~\citep{onebit2024,arb2025} or even sub-1-bit~\citep{stbllm2025}. Under such extreme bit-width compression, models often suffer from substantial performance degradation. 
Existing studies attribute this degradation primarily to numerical precision loss, and consequently focus on preserving the numerical precision of forward computation as much as possible~\cite{billm2024,arb2025}. A natural question then arises: \textit{Is the collapse of model performance solely caused by numerical precision in extremely low-bit quantization?}

\begin{figure}[tbp]
  \centering
  \includegraphics[width=0.45\columnwidth]{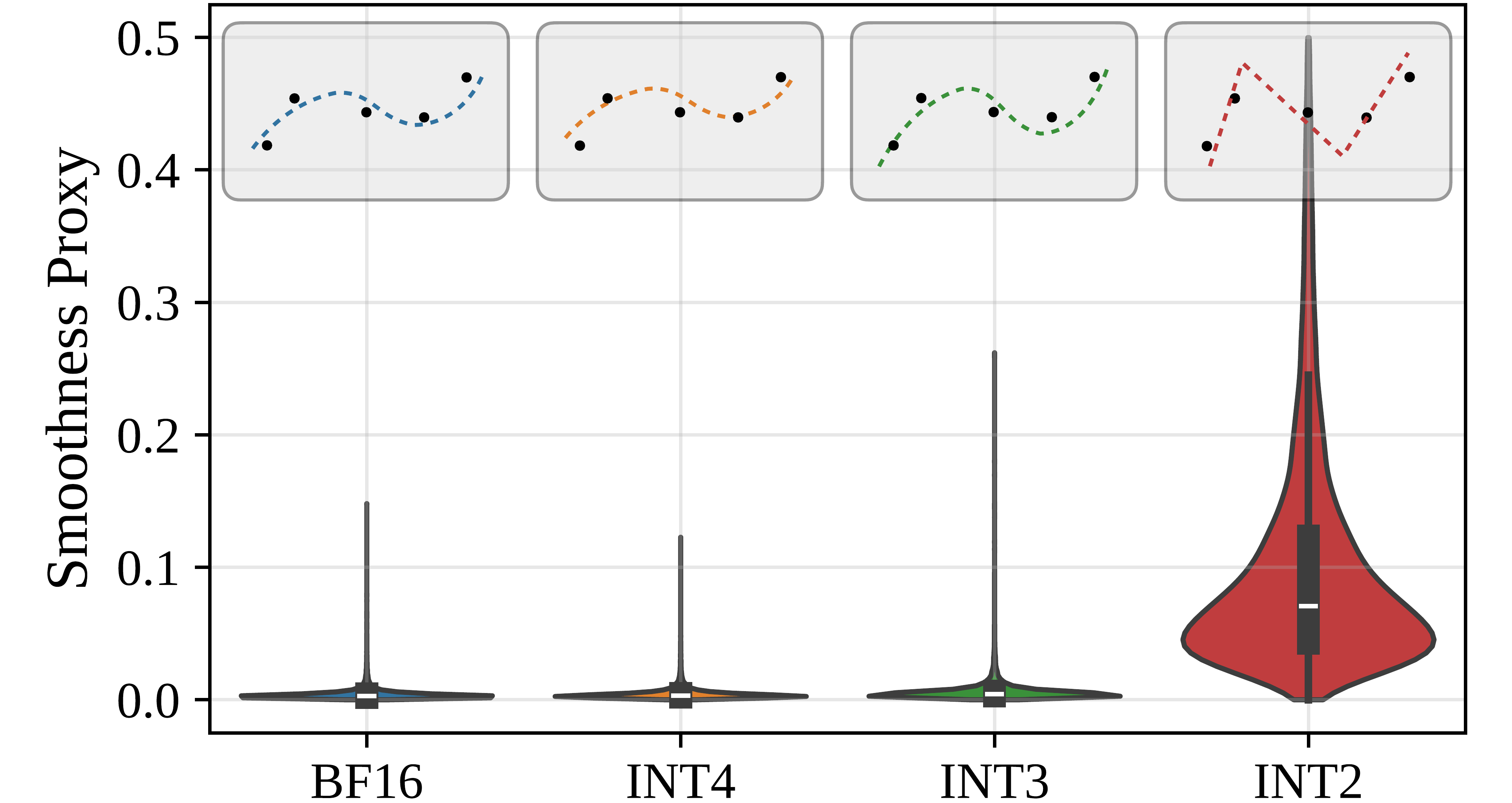}
  \caption{Smoothness degradation in extreme quantization. Smoothness score distributions of GPTQ-quantized LLaMA-2-7B under different bit-widths. Higher scores indicate worse smoothness.}
  \label{fig:intro}
\end{figure}

The answer may be ``no''. We empirically observe that such extremely quantized LLMs also suffer from pronounced smoothness degradation beyond fitting precision, which may constitute an additional source of capability loss, as shown in Figure~\ref{fig:intro}. Prior studies in machine learning have long connected smoothness to generalization, robustness, and training stability~\citep{sharpness2022,certified2019,understanding2022}. Models with poor smoothness are known to exhibit high sensitivity to small perturbations, resulting in unstable outputs and amplified generalization errors. Extensive evidence from classical machine learning and vision tasks further suggests that smoothness degradation is often accompanied by reduced performance and reliability~\citep{smooth2025,indirect2025,friendly2024,thinks2020}. Despite its well-established importance, however, smoothness remains largely underexplored in transformer-based LLMs, particularly in extreme quantization, motivating us to investigate its role in quantized LLM capability degradation.

Furthermore, by introducing a neighborhood model over text sequences together with reverse perplexity (rPPL), we theoretically show that the next-token probability distribution of quantized models collapses ``more rapidly'' than that of the original FP16 model. This phenomenon suggests that quantized models produce lower-quality next-token rankings at each decoding step, leaving a substantially narrower range of effective candidates for sampling. From a macroscopic perspective, this manifests as a significantly \textit{sparser effective decoding tree} generated by quantized models. More importantly, we find that this neighborhood collapse effect becomes increasingly severe as the quantization bit-width decreases. Additional analysis suggests that a direct way to mitigate such neighborhood collapse is to preserve model smoothness during quantization as much as possible, which aligns closely with our empirical observations.

Motivated by our empirical and theoretical findings, we propose simple strategies to mitigate smoothness degradation in extremely quantized LLMs. For post-training quantization (PTQ), we argue that existing methods rely on incomplete optimization objectives that focus mainly on reconstruction error while overlooking smoothness preservation. We therefore introduce learnable gradient preserving (\textbf{LGP}) to explicitly maintain original gradients during quantization. For quantization-aware training (QAT), we find that smoothness degradation mainly appears in the intermediate hidden-state gradients, and thus introduce a loss of gradient regularization (\textbf{LGR}) during training. Experiments on both PTQ and QAT show that smoothness yields additional performance gains. 

Rather than proposing a competing algorithm to others, we highlight smoothness preservation as a key design principle for extreme quantization. Our analysis suggests that in extremely low-bit settings, forward fitting and backward preservation are hard to optimize jointly, with the latter being more sensitive to bit-width reduction. Moreover, solution-space analysis shows that low-bit quantized weights capable of preserving both forward and backward behaviors do not disappear, but instead become increasingly narrow. Therefore, the objective of extreme quantization should no longer be to seek a lossless critical point on hidden-state reconstruction error or perplexity, but rather to achieve a principled trade-off between fitting precision and smoothness under limited bit-width budgets. Overall, this paper makes the following three contributions:

\begin{itemize}
    \item \textbf{Discovery.} We establish a feasible smoothness proxy for transformer-based LLMs, and reveal the smoothness degradation problem in extremely quantized LLMs. Moreover, through the sequence neighborhood modeling and rPPL, we uncover the decoding-tree sparsification effect caused by smoothness degradation.
    \item \textbf{Validation.} We design simple yet effective methods, including LGP for PTQ and LGR for QAT, to verify that smoothness enhancement brings positive performance gains to prediction distributions under extreme quantization.
    \item \textbf{Guidance.} We identify limitations in existing quantization objectives and provide an in-depth analysis of the feasibility and necessity of considering smoothness into extreme quantization. Our findings offer practical guidance for future algorithm design.
\end{itemize}

\section{Preliminary}

\subsection{Network Smoothness}

Lipschitzness is the most relevant metric of smoothness in neural networks. Generally, let $f: \mathcal{D} \subseteq \mathbb{R}^n \to \mathbb{R}^m$ be a function defined on a domain $\mathcal{D}$. The function $f$ is said to be $C$-Lipschitz continuous with respect to the $\alpha$-norm if there exists a constant $C > 0$ such that for all $\mathbf{x}, \mathbf{y} \in \mathcal{D}$:
\begin{equation}
    \|f(\mathbf{x}) - f(\mathbf{y})\|_\alpha \leq C \|\mathbf{x} - \mathbf{y}\|_\alpha.
\end{equation}
The Lipschitz constant can also be simply given by $C = \sup_{\mathbf{x} \in \mathcal{D}} \|\nabla_{\mathbf{x}} f\|_{\tilde{\alpha}}$,
where $\nabla_{\mathbf{x}} f$ is the Jacobian of $f$ with respect to the input $\mathbf{x}$. Here, $\tilde{\alpha}$ denotes the dual norm of $\alpha$ if $m=1$; otherwise, $\tilde{\alpha} = \alpha$. For simplicity, we fix $\alpha=\tilde{\alpha}=2$ in this paper. A smaller constant $C$ implies limited output variation under small input perturbations. It is closely associated with the generalization and robustness. Unfortunately, computing the exact value of $C$ is NP-hard~\citep{lipschitz2018}. Therefore, we approximately characterize the smoothness solely by estimating the upper and lower bounds of $C$.

The simplest, yet significantly \textit{loose}, upper bound is the product of the Lipschitz constants of each layer.
Moreover, the lower bound is estimated by sampling a small subset $S$ from the domain $\mathcal{D}$.
Consider an $L$-layer network $f_{\boldsymbol{\theta}}$ parameterized by $\boldsymbol{\theta}$, defined as $f_{\boldsymbol{\theta}} = f^{(L)} \circ f^{(L-1)} \circ \cdots \circ f^{(1)}$. An upper bound and lower bound on its Lipschitz constant is given by
\begin{equation}
    C_{\text{lower}} \leq \sup_{\mathbf{x} \in S} \|\nabla_{\mathbf{x}} f_{\boldsymbol{\theta}} \|_2 \leq C \leq \prod_{i=1}^L \sup_{\mathbf{x}^{(i-1)} \in \text{dom}(f^{(i)})} \|\nabla_{\mathbf{x}^{(i-1)}} f^{(i)}\|_2 = C_{\text{upper}},
\end{equation}
where $\mathbf{x}^{(i-1)}$ denotes the input for the $i$-th layer and $\text{dom}f$ is the definition area of $f$. $C_{\text{lower}}$ is more accurate and reduces the complexity of accurately estimating $C$. However, a drawback is that it merely reflects the local gradient magnitude at specific points rather than the global landscape. An alternative in computer version is to compute the \textbf{expected input gradient} $C_{\text{avg}} = \mathbb{E}_{\mathbf{x} \in S} \|\nabla_{\mathbf{x}} f_{\boldsymbol{\theta}} \|_2$.
Although this metric does not strictly satisfy the definition of Lipschitz constant, it provides a valid and often more practical estimate of $C$. For more details, please refer to \citet{some2024}.

\subsection{Model Quantization}

Model quantization converts the 16-bit floating-point formats commonly used in LLMs into low-bit representations~\citep{gptq2022,smooth2023}. While most studies focus on low-bit integer representations, floating-point formats~\citep{optimizing2024} and codebook-based encoding~\citep{quip2024,aqlm2024} are also investigated. In this work, we focus on fixed-point quantization for our analysis.

Quantization converts precision through scaling and shifting. The process is formulated as
\begin{equation}
    Q(w) = \text{clamp}(\lfloor \frac{w}{h} \rceil + z, 0, 2^N - 1),
\end{equation}
where $\lfloor \cdot \rceil$ denotes the rounding-to-nearest, and $\text{clamp}$ represents the clipping operation. The parameters $h$ and $z$ are obtained as follows:
\begin{equation}
\begin{aligned}
    h &= \frac{\max(w) - \min(w)}{2^N - 1},\\
    z &= -\lfloor \min(w) / h \rceil.
\end{aligned}
\end{equation}
Here, $N$ denotes the bit-width of the integer. For memory efficiency, matrix rows or columns are often quantized in groups, each sharing the same $h$ and $z$, typically of size 64 or 128.

Correspondingly, the dequantization process maps the discrete integer values back to the floating-point domain to enable subsequent computation. It can be defined as
\begin{equation}
    \hat{w}=Q^{-1}(Q(w))=h\cdot(Q(w)-z).
\end{equation}
The recovered value $\hat{w}$ serves as an approximation of the original weight $w$, with the quantization error determined by the bit-width $N$, the clipping range, and the group-wise statistics.

\begin{figure}[t]
  \centering
  \begin{subfigure}{0.32\columnwidth}
    \centering
    \includegraphics[width=\linewidth]{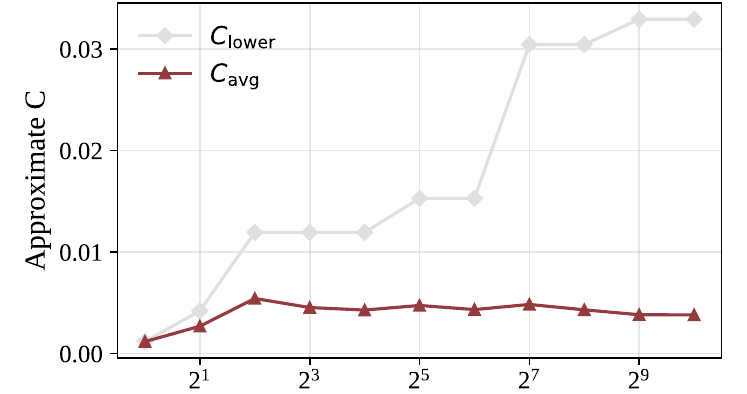}
    \caption{BF16 (Original)}
    \label{fig:proxy_a}
  \end{subfigure}
  \hfill
  \begin{subfigure}{0.32\columnwidth}
    \centering
    \includegraphics[width=\linewidth]{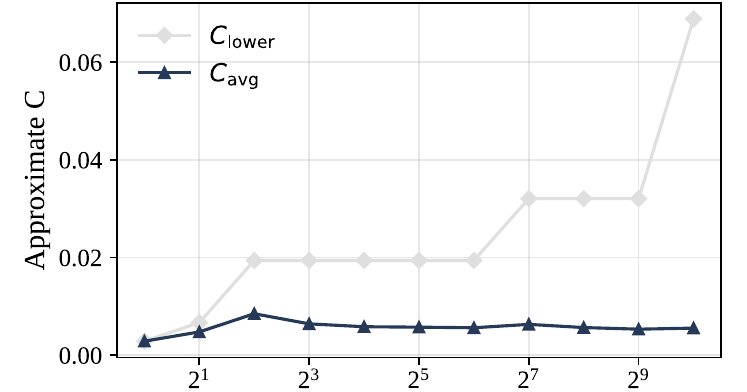}
    \caption{INT3 (GPTQ)}
    \label{fig:proxy_b}
  \end{subfigure}
  \hfill
  \begin{subfigure}{0.32\columnwidth}
    \centering
    \includegraphics[width=\linewidth]{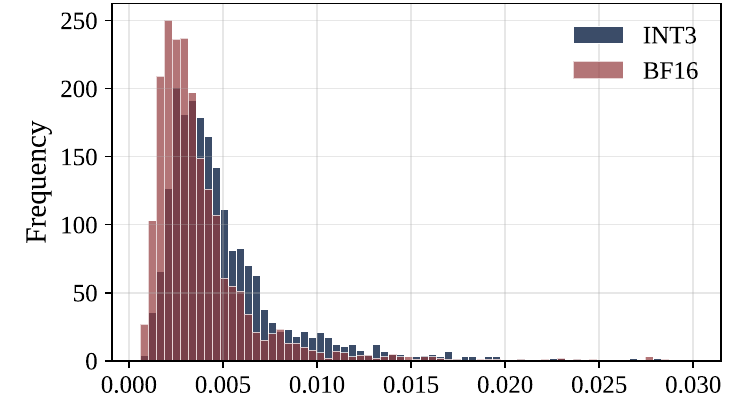}
    \caption{Input gradient distribution}
    \label{fig:proxy_c}
  \end{subfigure}
  \caption{Approximate Lipschitz constant, i.e. expected input gradient $C_{\text{avg}}$, of LLaMA-2-7B on one input sequence under different precisions.}
  \label{fig:proxy}
\end{figure}

\section{Empirics: Proxy and Smoothness}

While Lipschitz constant bounds and expected input gradient effectively characterize model smoothness, existing conclusions are predominantly derived from simple architectures or vision tasks, such as MLPs~\citep{efficiently2022}, ResNets~\citep{thinks2020}, or ViTs~\citep{some2024}. It remains unclear whether these findings hold for transformer-based LLMs and quantized LLMs. In this section, we discuss the smoothness degradation problem in quantized LLMs under the premise of validating the applicability of the proxy.

$C_{\text{avg}}$ is a differentiable metric, yet it deviates from the strict definition of $C$. We now need to address two pivotal questions: \textbf{(a) Can $C_{\text{avg}}$ approximate $C$ in transformer-based LLMs?} and \textbf{(b) Does this approximation hold for quantized LLMs?}

To facilitate measuring the impact of input perturbations on LLM outputs, we define $f$ as the language modeling objective combined with the cross-entropy loss. Consequently, $f$ takes the form $f: \mathbb{R}^n \to \mathbb{R}$. A key benefit is that $\nabla_{\mathbf{x}} f$ simplifies to a gradient vector instead of a Jacobian matrix, thereby avoiding the excessive computation and memory overhead associated with Jacobians.

We provide a formal definition of $\nabla_{\mathbf{x}} f$ in the context of LLMs. For an input sequence $(w_1, \dots, w_T)$, let $\mathbf{x}_t^{(i)}$ denote the input hidden state of token $w_t$ at layer $i = 1, \dots, L$. The smoothness proxy, referred to as ``input gradient'', is defined as $\nabla_{\mathbf{x}^{(0)}} f$, where $\mathbf{x}^{(0)}$ is the token embedding of $w_t$.

Since $C_{\text{lower}}$ is shown to be a reasonably accurate estimate of the Lipschitz constant~\citep{some2024}, we investigate the relationship between $C_{\text{avg}}$ and $C_{\text{lower}}$ on LLaMA-2-7B. As shown in Figure~\ref{fig:proxy_a}, $C_{\text{lower}}$ converges as the number of input tokens increases. Although $C_{\text{lower}}$ exhibits a few distinct spikes, these are attributed to a very small number of tokens with large gradients. As illustrated in Figure~\ref{fig:proxy_c}, the gradients for nearly all tokens remain below 0.02. In the absence of these outliers, the distribution of $C_{\text{lower}}$ would be flat. Meanwhile, despite a magnitude gap between the two, $C_{\text{avg}}$ exhibits a trend similar to that of $C_{\text{lower}}$. Therefore, although $C_{\text{avg}}$ does not strictly adhere to the definition of $C$, it remains a valid metric for estimating trends in LLM smoothness.

This conclusion also holds for the quantized models shown in Figure~\ref{fig:proxy_b}. Additionally, Figure~\ref{fig:proxy_c} illustrates the shift in input gradients for quantized models. We observe that at 3-bit quantization, the input gradients exhibit a noticeable increase, even though the quantization loss remains minimal. Figure~\ref{fig:intro} and~\ref{fig:proxy_c} clearly show that \textit{there exists a metric independent of the forward fitting accuracy in quantization, the smoothness captured by the input-gradient, which degrades sharply as the quantization bit-width decreases.}

\section{Theory: Sequence Neighborhood Modeling}

Building on the empirical findings, we further investigate from a theoretical perspective how the behavior of quantized models changes. A widely adopted approach for evaluating quantized model capability is to evaluate perplexity on a validation dataset. This metric essentially reflects how close the predicted probability $p(w_{T+1}|w_{1:T})$ of the next token $w_{T+1}$ is to 1, given a context $w_{1:T}$ from the dataset. However, this evaluation paradigm overlooks a fundamental fact: in practical usage, LLMs rely on top-k or top-p sampling, where multiple candidate tokens beyond $w_{T+1}$ are required to be assigned relative high probabilities.

\begin{figure}[t]
  \centering
  \begin{subfigure}{0.28\columnwidth}
    \centering
    \includegraphics[width=\linewidth]{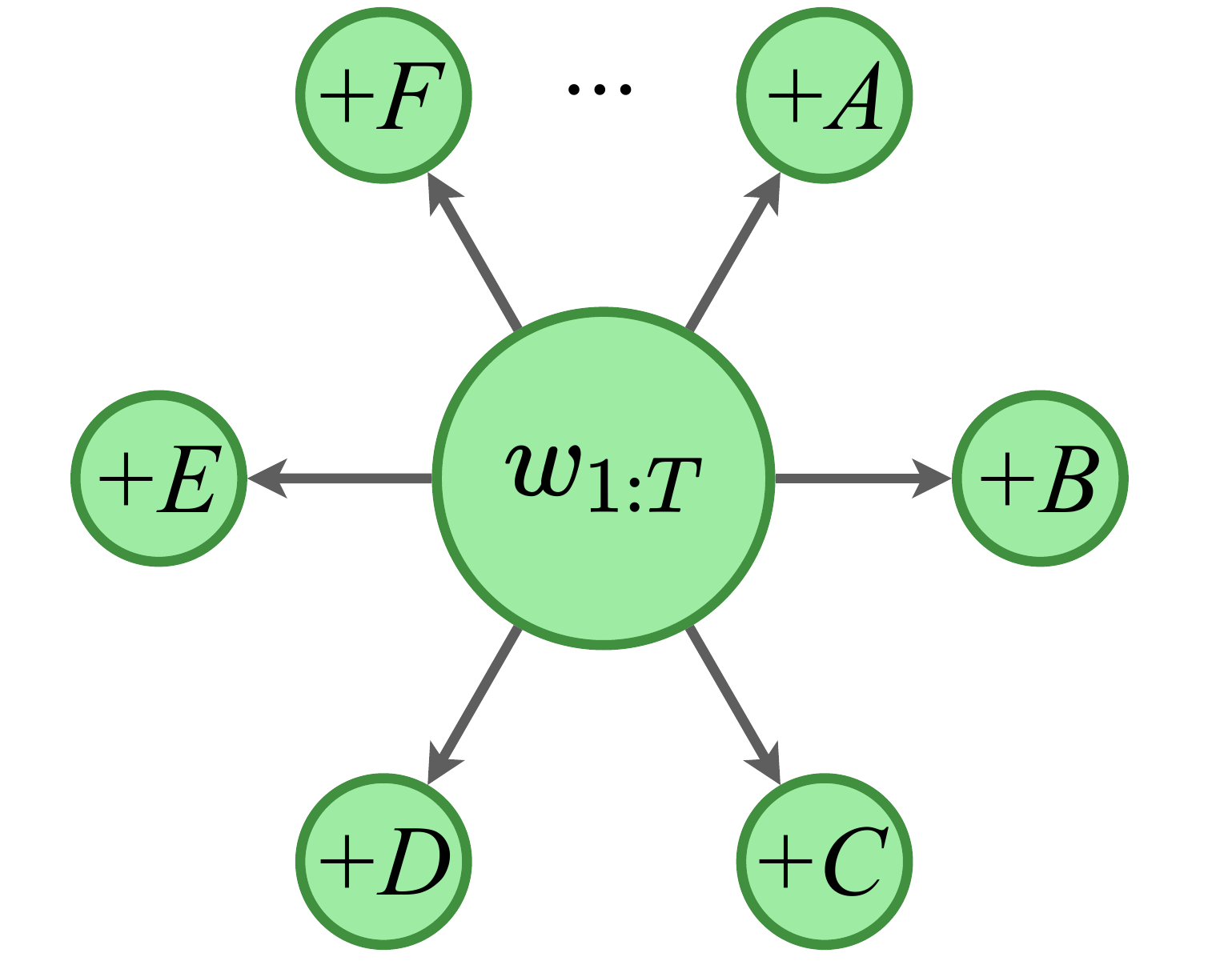}
    \caption{Definition of $N_\delta(w_{1:T})$}
    \label{fig:neighbor}
  \end{subfigure}
  \hspace{10mm}
  \begin{subfigure}{0.48\columnwidth}
    \centering
    \includegraphics[width=\linewidth]{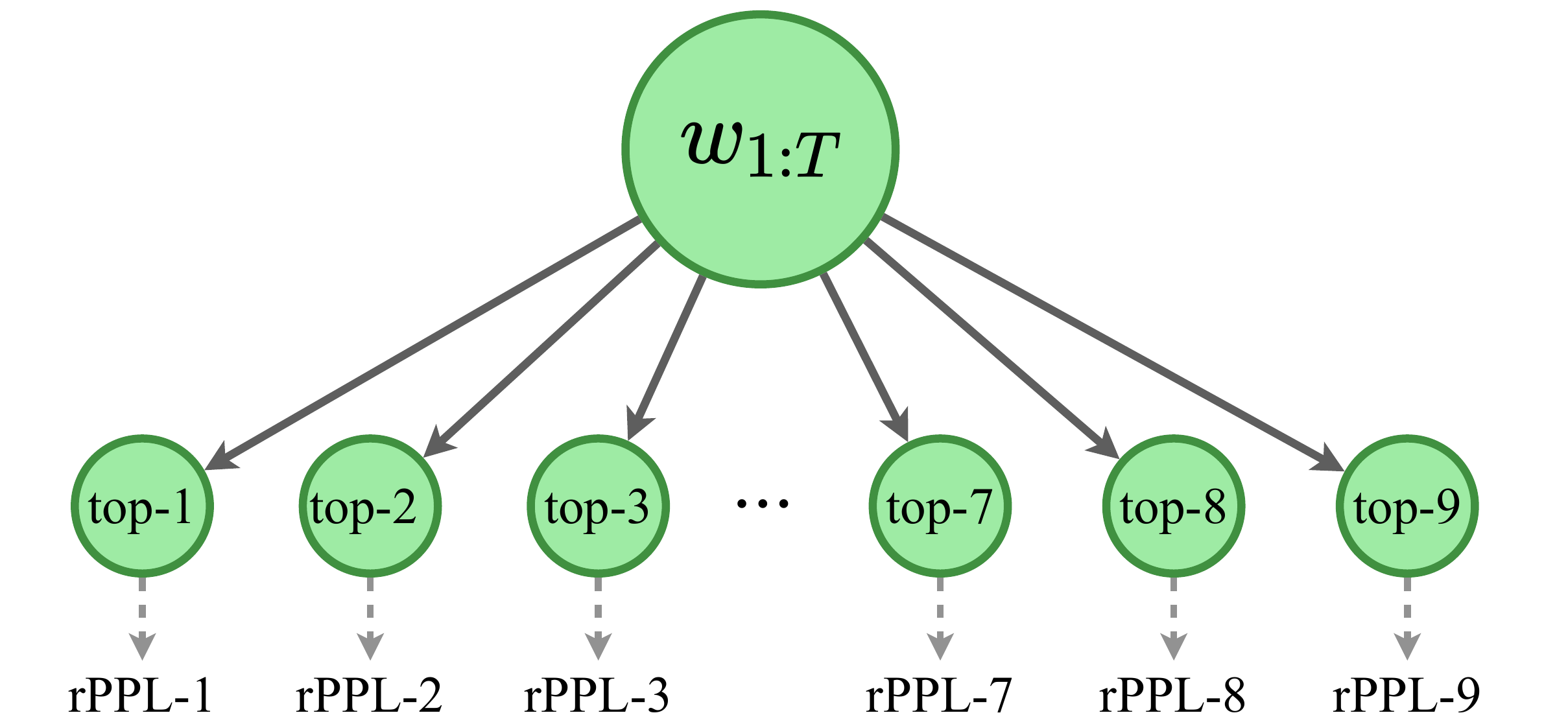}
    \caption{Definition of rPPL-$k$}
    \label{fig:rppl}
  \end{subfigure}
  \caption{Key definitions in sequence neighborhood modeling.}
  \label{fig:def}
\end{figure}

The ability of an autoregressive language model are reflected in step-by-step decoding. We define a context $c = w_{1:T}$ as a point in the sequence space, and view the newly added token $w$ as a small perturbation $\delta=1$. Then the $\delta$-neighborhood of the sequence space centered at $c$ can be defined as:
\begin{equation}
    N_\delta(c) = \{ c \cup \{w\} \mid w \in |\mathcal{V}| \},
\end{equation}
where $\mathcal{V}$ denotes the model vocabulary, as shown in Figure~\ref{fig:neighbor}. The function $f$, i.e., language modeling combined with cross-entropy loss, is defined over the above sequence space and maps a sequence to a real-valued score. The directional derivative of $f$ at point $c$ in direction $w$ is defined as:
\begin{equation}
   \nabla_{c\rightarrow c+w}f = f(c+w)-f(c).
\end{equation}
Since $f(c)$ is a constant, we redefine the directional derivative as $\nabla_{c\rightarrow c+w}f = f(c+w)$. $\nabla_{c\rightarrow c+w}f$ measures the quality of perturbations of $c$ along different directions: if the perturbation leads to worse sequence quality, $\nabla_{c\rightarrow c+w}f$ will be larger.

The decoding of an LLM can be viewed as a sequence perturbation process starting from a given point. For a context $c=w_{1:T}$, a quantized model $\mathcal{M}_Q$ determines possible directions of the next perturbation by computing $p(\cdot | w_{1:T})$. To evaluate the quality of such perturbations, we compute the perplexity of $c+w$ using the original full-precision model $\mathcal{M}_{FP16}$, a metric we refer to as \textbf{reverse perplexity} (rPPL). We denote the perplexity computed by perturbing $c$ along the direction of the token ranked $k$-th by $p(\cdot | w_{1:T})$ as rPPL-$k$, as shown in Figure~\ref{fig:rppl}. In particular, perturbing along the most probable token (greedy decoding) direction yields rPPL-1. We do not compute perplexity over all vocabulary directions. Instead, we focus on the most commonly feasible sampling range, such as perturbations corresponding to the top-40 predicted tokens.

\begin{figure}[ht]
  \centering
  \begin{subfigure}{0.40\columnwidth}
    \centering
    \includegraphics[width=\linewidth]{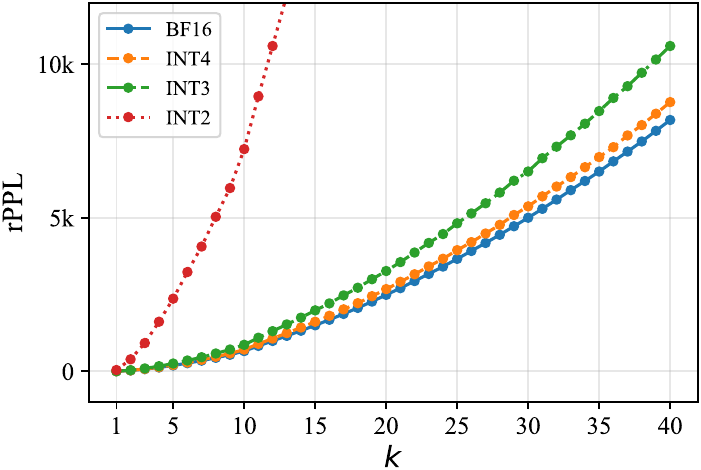}
    \caption{Quantized by GPTQ}
    \label{fig:rppl_gptq}
  \end{subfigure}
  \hspace{10mm}
  \begin{subfigure}{0.40\columnwidth}
    \centering
    \includegraphics[width=\linewidth]{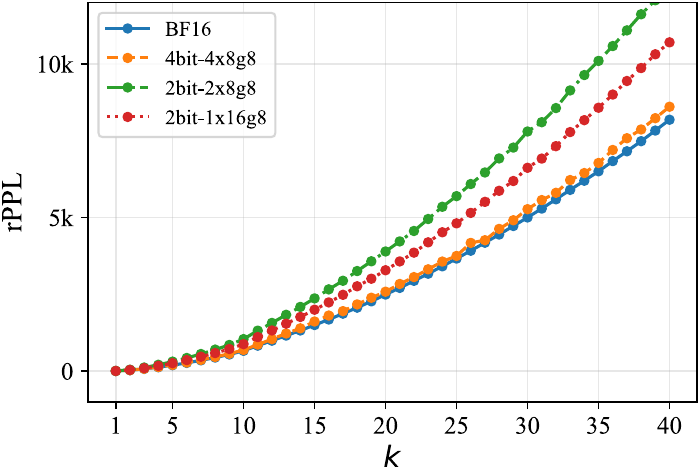}
    \caption{Quantized by AQLM}
    \label{fig:rppl_aqlm}
  \end{subfigure}
  \caption{Results of rPPL-1 to rPPL-40 on original and quantized LLaMA-2-7B models. For GPTQ, we evaluate 4-bit, 3-bit, and 2-bit quantization with group size 128. For the vector-quantization method AQLM, we compare codebook settings of 4$\times$8, 2$\times$8, and 1$\times$16 with group size 8.}
  \label{fig:rppls}
\end{figure}

rPPL reflects the quality of different perturbation directions guided by quantized models. We compare rPPL results on C4 on different models and quantization algorithms, as shown in Figure~\ref{fig:rppls}. For both the original and quantized models, rPPL-$k$ increases as k becomes larger, indicating that only top-ranked predictions are effective in each decoding step, which aligns with intuition. However, our focus is neither the absolute value of rPPL nor this monotonically increasing trend itself. More importantly, the growth becomes substantially steeper and faster as the quantization bit-width decreases. This rapid collapse behavior shows little correlation with the starting point rPPL-1, which mainly reflects the quality of single-point predictions.

\begin{figure}[t]
  \centering
  \begin{subfigure}{0.33\columnwidth}
    \centering
    \includegraphics[width=\linewidth]{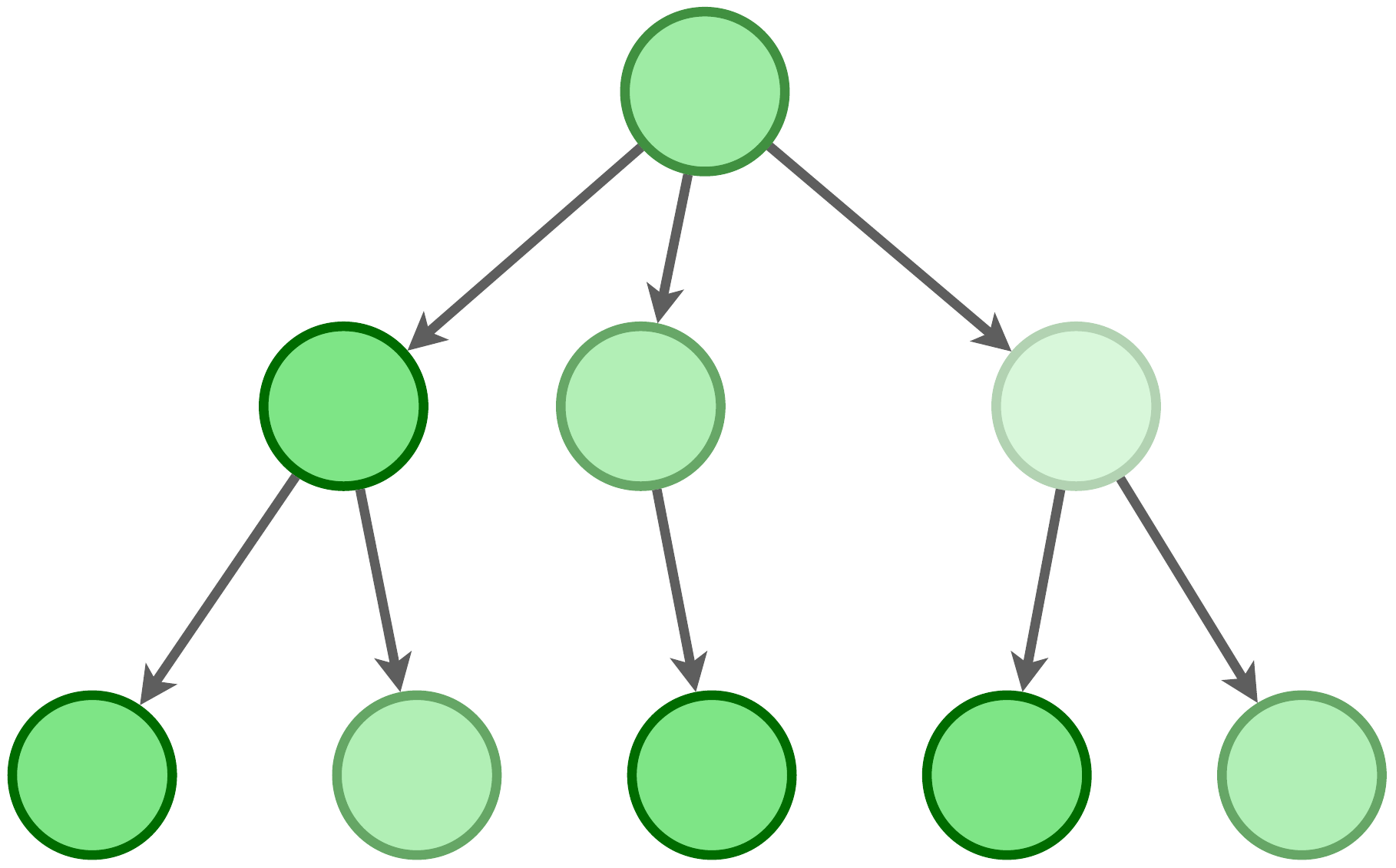}
    \caption{Decoding tree of FP16 model}
    \label{fig:tree_fp16}
  \end{subfigure}
  \hspace{10mm}
  \begin{subfigure}{0.33\columnwidth}
    \centering
    \includegraphics[width=\linewidth]{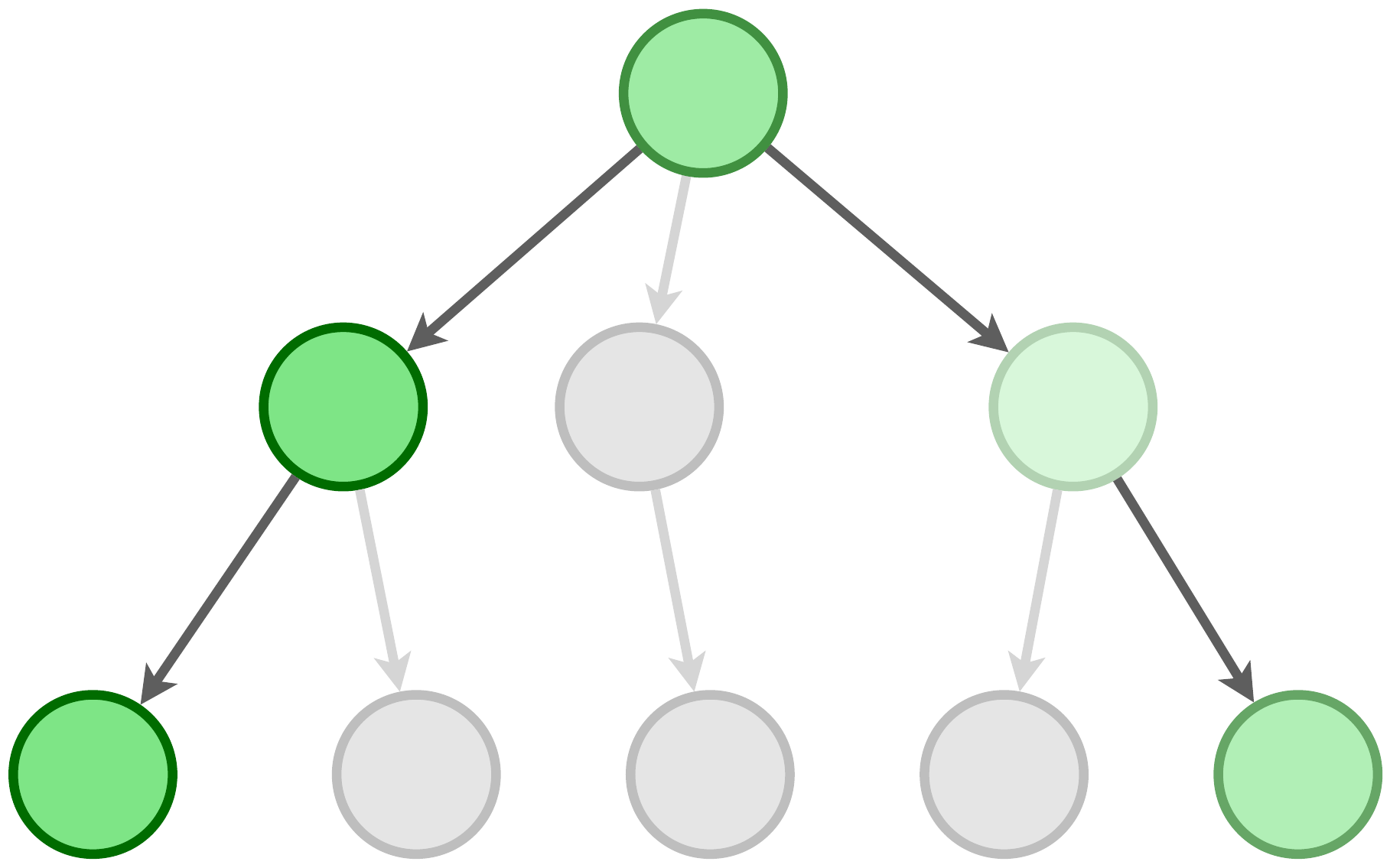}
    \caption{Decoding tree of INT2 model}
    \label{fig:tree_quant}
  \end{subfigure}
  \caption{Illustrative decoding trees of original and quantized models. For any given node, darker child nodes indicate higher predictive quality in that direction, while gray nodes represent near-invalid tokens with extremely poor quality.}
  \label{fig:tree}
\end{figure}

The lower the quantization bit-width, the faster the rPPL collapse becomes. This implies that as quantization approaches the extreme regime, more perturbation directions around any point in the sequence space become ineffective under the guidance of the quantized model. We illustrate this intuition together with Figure~\ref{fig:tree}. After quantization, the number of effective child nodes at each node of the decoding tree decreases, and this effect becomes increasingly severe at lower bit-widths. While the original FP16 model may provide around 10 effective predictions, the number of effective tokens in a quantized model may shrink to 5 or even fewer. Such local shrinkage naturally accumulates as decoding proceeds. A direct consequence is that the decoding tree becomes progressively sparser. Therefore, the closer quantization moves toward the extreme low-bit regime, the sparser the decoding tree becomes, and the less likely the model is to decode high-quality generation paths. Importantly, this phenomenon cannot be revealed by conventional perplexity evaluation, since perplexity only measures the likelihood of a single path within the decoding tree.

rPPL can be viewed as the directional derivative of a quantized model along different prediction directions. Figure~\ref{fig:rppls} shows that models with lower quantization bit-width exhibit much more drastic variation in directional derivatives around points in the sequence space. This variation can be expanded as:
\begin{equation}
    \nabla_{c\rightarrow c+w}f = f(c+w)-f(c)\approx \nabla_c f^\top w + \frac{1}{2}w^\top H w.
\end{equation}
Ignoring the second-order term H, we obtain $\nabla_{c\rightarrow c+w}f = \nabla_c f^\top w$. Here, $\nabla_{c\rightarrow c+w}f$ and $\nabla_c f$ are different quantities: the former denotes the directional derivative of $f$ at point $c$ along a specific direction $w$, while the latter denotes the gradient of $f$ with respect to the input $c$. To suppress the rapid growth of $\nabla_{c\rightarrow c+w}f$ caused by quantization, we can optimize $\nabla_c f^\top w$. Since $\|\nabla_c f^\top w\|\leq \|\nabla_c f\|\cdot\|w\|$, and $\|w\|$ depends on the specific token direction and cannot be controlled, the practical way to minimize the above expression is to keep $\|\nabla_c f\|$ as stable or as small as possible during quantization. This not only explains our previous empirical observations, but also provides a feasible direction for optimizing extreme quantization algorithms.

\section{Simple Smoothness Preservation Method}

In this section, we propose simple smoothness-preserving methods for both PTQ and QAT to examine whether smoothness brings performance gains to quantized models. Due to space limitations, we provide all experimental settings and ablations in Appendix~\ref{subsec:setup} and~\ref{subsec:ablation}.

\subsection{Learnable Gradient Preservation for PTQ}
\label{subsec:lgp}

\begin{figure*}[t]
\centering
  \begin{minipage}{0.45\textwidth}
  \centering
  \includegraphics[width=0.7\linewidth]{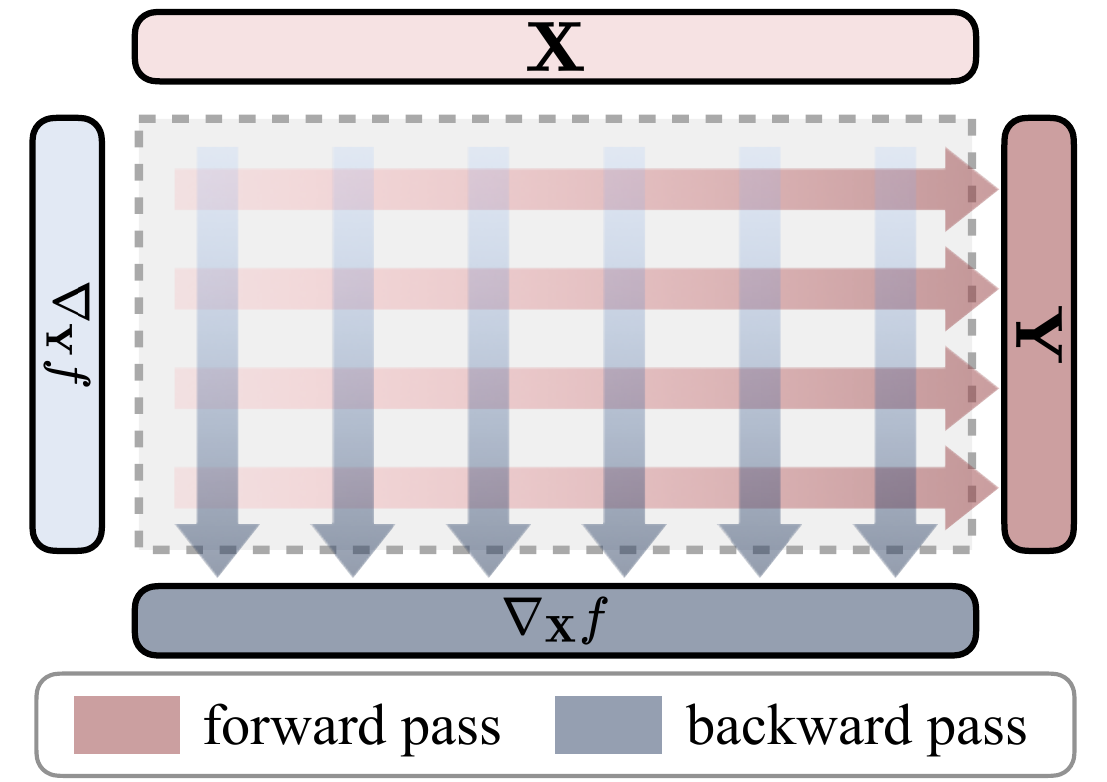}
  \caption{The row-wise forward pass and the column-wise backward pass.}
  \label{fig:method}
  \end{minipage}
  \hfill
  \begin{minipage}{0.50\textwidth}
  \centering
  \includegraphics[width=0.8\linewidth]{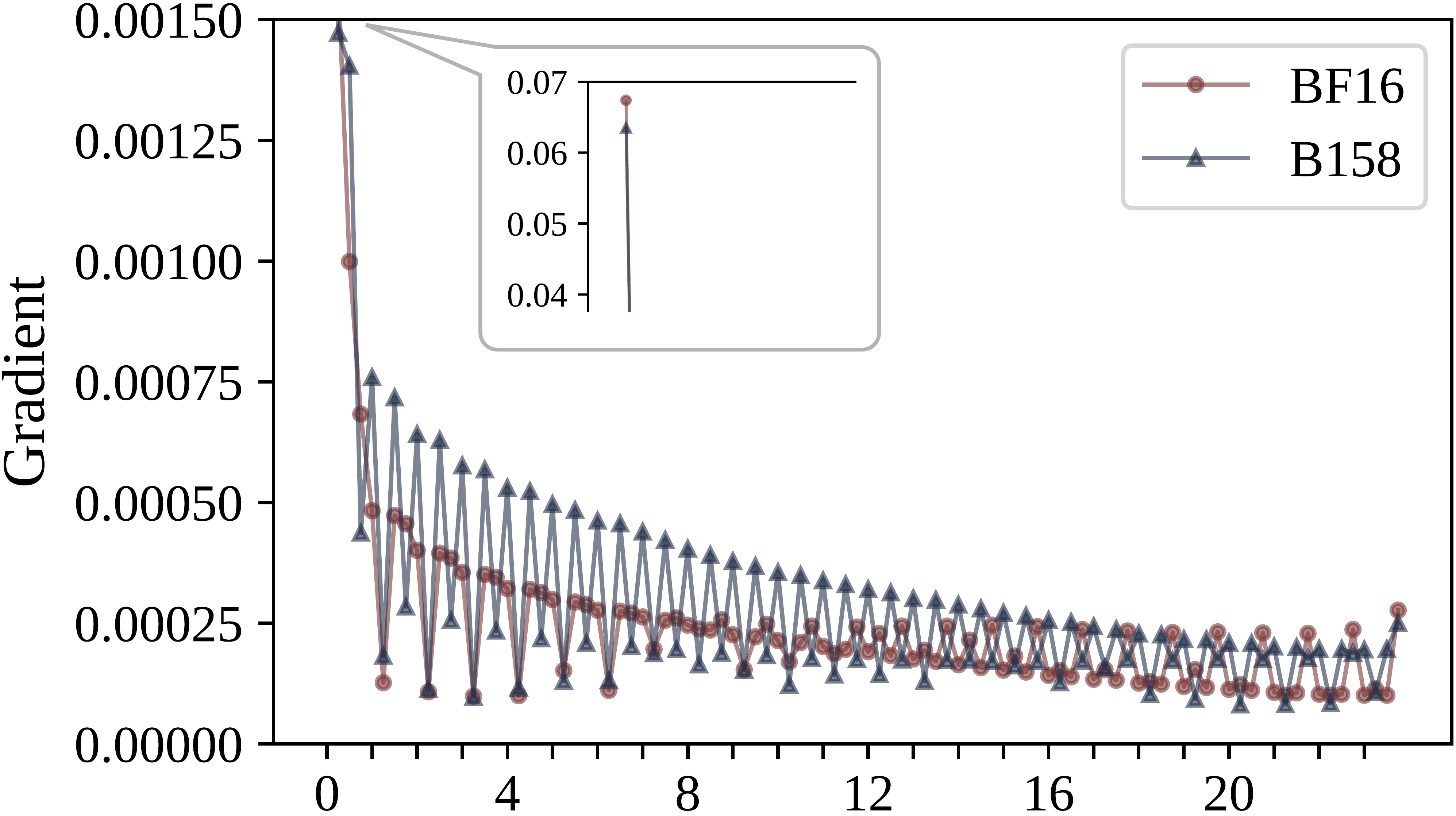}
  \caption{Layer-wise gradient of BF16 and B158 models. The spike denotes ``gradient ridge''.}
  \label{fig:layer_grad}
  \end{minipage}
\end{figure*}

PTQ quantizes LLMs in a layer-wise manner by aligning the output activations of quantized modules with those of the original model. For a linear layer $\mathbf{Y}=\mathbf{WX}$, existing PTQ methods mainly optimize the forward reconstruction objective $\|\mathbf{WX}-\hat{\mathbf{W}}\mathbf{X}\|_F^2$, which preserves forward fitting accuracy but does not guarantee backward gradient preservation after quantization. Considering backward propagation, let $\nabla_{\mathbf{Y}}f=\mathbf{G}$ and the input gradient is given by $\nabla_{\mathbf{X}}f=\mathbf{W}^\top\mathbf{G}$, indicating that smoothness preservation additionally requires maintaining gradient propagation, i.e., minimizing $\|\mathbf{W}^\top\mathbf{G}-\hat{\mathbf{W}}^\top\mathbf{G}\|_F^2$. Therefore, preserving both model accuracy and smoothness requires jointly optimizing forward fitting and backward preservation:
\begin{equation}
    \min_{\hat{\mathbf{W}}} \underbrace{\|\mathbf{WX} - \hat{\mathbf{W}}\mathbf{X}\|^2_F}_{\text{accuracy}} + \underbrace{\|\mathbf{W}^\top \mathbf{G} - \hat{\mathbf{W}}^\top \mathbf{G} \|^2_F}_{\text{smoothness}}.
\label{eq:core}
\end{equation}
As illustrated in Figure~\ref{fig:method}, these two objectives are largely orthogonal, suggesting that classical PTQ objectives are inherently incomplete. 


The objective in $\|\mathbf{WX}-\hat{\mathbf{W}}\mathbf{X}\|_F^2$ admits a high-precision approximate closed-form solution, and methods such as GPTQ exploit second-order information (i.e., the Hessian $\mathbf{H}=\mathbf{XX}^\top$) to iteratively quantize weights while compensating for induced errors, forming the backbone of many PTQ approaches. However, the joint smoothness objective in Equation~\ref{eq:core} is incompatible with such GPTQ-style solvers due to orthogonal optimization structures: the forward reconstruction term follows a row-wise update scheme, whereas smoothness preservation in $\|\mathbf{W}^\top\mathbf{G}-\hat{\mathbf{W}}^\top\mathbf{G}\|_F^2$ requires column-wise consistency in gradient propagation. To alleviate this issue, we select to build upon OmniQuant~\citep{omniquant2023}, which introduces learnable weight clipping parameters $\gamma$ and $\beta$,
\begin{equation}
    h = \frac{\gamma\max(w) - \beta\min(w)}{2^N - 1}, \quad
z = -\lfloor \beta\min(w) / h \rceil,
\end{equation}
and optimizes them via layer-wise distillation. Based on this framework, we introduce learnable gradient preservation (LGP) to explicitly maintain smoothness, yielding the joint objective:
\begin{equation}
\min_{\hat{\boldsymbol{\theta}}} \|z_{\boldsymbol{\theta}} - z_{\hat{\boldsymbol{\theta}}}\|_F^2 + \alpha_1 \|\nabla_{\mathbf{X}} f_{z_{\boldsymbol{\theta}}} - \nabla_{\mathbf{X}} f_{z_{\hat{\boldsymbol{\theta}}}}\|_F^2,
\end{equation}
where $\alpha_1$ balances fitting and smoothness. This formulation helps effectively preserve smoothness in quantization, even though exact closed-form solutions for the joint objective remain an open problem.

\begin{table*}[t!]
\centering
\small
\resizebox{\textwidth}{!}{ %
\begin{tabular}{@{\hspace{0.20cm}}c@{\hspace{0.20cm}}|c@{\hspace{0.3cm}}c|c@{\hspace{0.30cm}}c@{\hspace{0.40cm}}c@{\hspace{0.50cm}}c@{\hspace{0.50cm}}c@{\hspace{0.40cm}}c@{\hspace{0.30cm}}c@{\hspace{0.30cm}}c@{\hspace{0.30cm}}c@{\hspace{0.30cm}}c}
\toprule
Method & Wiki2 & C4 & BoolQ & OBQA & RTE & Wino. & Hella. & PIQA & MathQA & ARC-e & ARC-c & Avg. \\
\midrule
\multicolumn{13}{c}{\textbf{LLaMA-2-7B}\quad($\alpha_1 = 1e6$ in LGP)}\\
\cmidrule(lr){1-13}
BF16 & 5.47 & 6.97 & 77.71 & 44.20 & 62.82 & 69.14 & 76.00 & 79.11 & 28.31 & 74.54 & 46.25 & 62.01 \\
\cmidrule(lr){1-13}
GPTQ & 39.58 & 31.19 & 58.34 & 26.80 & 54.51 & 52.49 & 37.13 & 58.54 & 22.68 & 38.43 & 25.26 & 41.58 \\
OmniQ & \textbf{13.76} & 14.43 & \textbf{63.27} & 31.20 & 54.51 & \textbf{55.88} & 50.70 & 66.59 & 22.71 & 43.60 & 25.77 & 46.03 \\
+LGP & 13.77 & \textbf{14.37} & 62.35 & \textbf{33.00} & \textbf{55.23} & 55.25 & \textbf{50.81} & \textbf{67.95} & \textbf{23.38} & \textbf{45.45} & \textbf{26.62} & \textbf{46.67} \\
\midrule
\multicolumn{13}{c}{\textbf{Llama-2-13B}\quad($\alpha_1 = 1e5$ in LGP)}\\
\cmidrule(lr){1-13}
BF16 & 4.88 & 6.47 & 80.55 & 45.20 & 65.34 & 72.14 & 79.38 & 80.52 & 31.86 & 77.44 & 49.06 & 64.61 \\
\cmidrule(lr){1-13}
GPTQ & 30.79 & 29.81 & 54.19 & 27.60 & 50.18 & 51.14 & 41.08 & 61.37 & 22.24 & 38.64 & 26.02 & 41.38 \\
OmniQ & 9.30 & 10.90 & 66.69 & 33.20 & 55.23 & 58.17 & \textbf{60.38} & \textbf{71.82} & \textbf{25.90} & 57.15 & 31.83 & 51.15 \\
+LGP & \textbf{9.25} & \textbf{10.87} & \textbf{67.09} & \textbf{34.20} & \textbf{56.68} & \textbf{58.17} & 60.00 & 71.71 & 25.06 & \textbf{58.08} & \textbf{32.59} & \textbf{51.51} \\
\midrule
\multicolumn{13}{c}{\textbf{Qwen-3-0.6B}\quad($\alpha_1 = 1e4$ in LGP)}\\
\cmidrule(lr){1-13}
BF16 & 20.96 & 25.43 & 63.82 & 31.40 & 53.79 & 56.43 & 47.30 & 67.25 & 31.46 & 55.93 & 33.70 & 49.01 \\
\cmidrule(lr){1-13}
GPTQ & 7.1e3 & 2.4e4 & 41.99 & 23.00 & 51.62 & 50.83 & 25.59 & 52.56 & 19.39 & 25.84 & 25.00 & 35.09 \\
OmniQ & 310.2 & 141.8 & 61.07 & 25.00 & 52.71 & 48.78 & 28.28 & \textbf{56.96} & \textbf{22.81} & 31.86 & 22.26 & 38.86 \\
+LGP & \textbf{295.0} & \textbf{140.1} & \textbf{61.50} & \textbf{25.00} & \textbf{52.71} & \textbf{51.62} & \textbf{28.44} & 56.75 & 22.45 & \textbf{31.94} & \textbf{22.53} & \textbf{39.22} \\
\bottomrule
\end{tabular}
}
\caption{Main W2A16 quantization results of the evaluation experiment. The best scores are in bold.}
\label{tab:main}
\end{table*}

Table~\ref{tab:main} presents the evaluation results of incorporating LGP into PTQ. Compared to GPTQ, OmniQuant achieves substantial accuracy gains in 2-bit weight quantization, although a noticeable performance gap remains relative to the FP16 baseline. After integrating LGP, we first observe that the language modeling capability is maintained or even slightly enhanced. Furthermore, LGP demonstrates improvements across the majority of accuracy tasks for each model. Notably, although LGP does not explicitly optimize for fitting accuracy, it yields further improvements in both PPL and Accuracy. We attribute this success to the inherent advantages of smoothness, which mitigates erratic output shifts. We deeply discuss the reasons behind this change later in the paper.

\subsection{Loss of Gradient Regularization for QAT}
\label{subsec:lgr}

QAT typically relies on the language modeling loss and therefore also overlooks smoothness preservation. While one may expect $\nabla_{\mathbf{x}^{(0)}}f$ to serve as a direct constraint on model smoothness, we find that it is ineffective due to the existence of a \textit{gradient ridge} at the 0-th layer, as shown in Figure~\ref{fig:layer_grad}. To investigate this phenomenon, we compare layer-wise gradients between a BF16 model and its 1.58-bit counterpart~\citep{era2024}, monitoring input and output gradients at the \texttt{input\_layernorm} and \texttt{post\_attention\_layernorm} across layers. We observe that the quantized model exhibits significantly larger intermediate gradients, indicating reduced smoothness in hidden states, particularly in early and middle layers. However, the input gradients at the 0-th layer remain consistently high and nearly identical across both models, suggesting that this behavior is not induced by quantization. We hypothesize that this stems from the nature of embedding inputs, which lack semantic structure and lead to sparse representations, causing optimization to concentrate on ridge-like regions in the latent space rather than smooth valleys. We refer to this phenomenon as the ``Gradient Ridge'', an intrinsic property independent of weight quantization, which explains why $\nabla_{\mathbf{x}^{(0)}}f$ cannot reliably reflect smoothness degradation.

\begin{table*}[t!]
\centering
\small
\resizebox{\textwidth}{!}{ %
\begin{tabular}{c@{\hspace{0.20cm}}|@{\hspace{0.20cm}}c@{\hspace{0.20cm}}|c@{\hspace{0.6cm}}c|@{\hspace{0.30cm}}c@{\hspace{0.50cm}}c@{\hspace{0.50cm}}c@{\hspace{0.50cm}}c@{\hspace{0.40cm}}c@{\hspace{0.40cm}}c@{\hspace{0.40cm}}c@{\hspace{0.40cm}}c@{\hspace{0.30cm}}c@{\hspace{0.30cm}}c}
\toprule
Size & Method & Wiki2 & C4 & Wino. & Hella. & PIQA & BoolQ & OBQA & ARC-e & ARC-c & Avg. \\
\midrule
\multirow{3}{*}[-0ex]{0.4B} & BF16 & 58.85 & 67.76 & 50.75 & 26.85 & 53.97 & 61.77 & 26.40 & 31.48 & 23.04 & 39.18 \\
& B158 & 62.86 & \textbf{68.11} & 49.80 & 25.51 & 53.21 & 62.05 & \textbf{25.00} & \textbf{30.64} & 23.46 & 38.52 \\
& +LGR & \textbf{62.28} & 69.44 & \textbf{50.04} & \textbf{26.22} & \textbf{53.54} & \textbf{62.05} & 24.40 & 29.76 & \textbf{24.49} & \textbf{38.64} \\
\midrule
\multirow{3}{*}[-0ex]{1.0B} & BF16 & 60.29 & 72.43 & 46.88 & 27.13 & 55.28 & 42.29 & 24.00 & 30.93 & 23.21 & 35.67 \\
& B158 & \textbf{65.52} & 77.12 & 48.30 & 25.70 & 49.62 & 37.83 & 26.20 & 27.15 & 24.57 & 34.20 \\
& +LGR & 66.97 & \textbf{76.15} & \textbf{50.28} & \textbf{25.75} & \textbf{50.54} & \textbf{37.83} & \textbf{28.20} & \textbf{27.61} & \textbf{25.43} & \textbf{35.09} \\
\midrule
\multirow{3}{*}[-0ex]{1.7B} & BF16 & 57.76 & 65.87 & 49.41 & 27.07 & 56.04 & 39.36 & 24.20 & 32.24 & 23.81 & 36.02 \\
& B158 & \textbf{55.61} & 70.87 & \textbf{51.22} & 26.07 & 49.62 & 37.83 & 26.40 & 26.30 & \textbf{25.77} & 34.74 \\
& +LGR & 55.80 & \textbf{68.22} & 50.51 & \textbf{26.17} & \textbf{51.03} & \textbf{38.41} & \textbf{26.80} & \textbf{27.02} & 24.32 & \textbf{34.89} \\
\bottomrule
\end{tabular}
}
\caption{Main quantization results on our trained FP16 and B158 models. The best scores are in bold.}
\label{tab:main_qat}
\end{table*}

To enable extremely low-bit models to automatically acquire smoothness during training, we propose a gradient regularization loss (LGR). For the existence of the gradient ridge, we avoid using 0-th layer inputs and instead apply regularization on the 1-st layer hidden states. Specifically, we define the smoothness objective as $\mathcal{L}_{\text{smooth}} = \frac{1}{N}\sum_{i=1}^N \|\nabla_{\mathbf{x}_i^{(1)}} f\|_F^2$, where $N$ is the sequence length. The overall training objective is then given by $\mathcal{L} = \mathcal{L}_{\text{lm}} + \alpha_2 \mathcal{L}_{\text{smooth}}$, with $\alpha_2$ controlling the strength of smoothness regularization. This formulation allows smoothness to be explicitly encouraged in extremely low-bit training while avoiding the \textit{instability} introduced by 0-th layer gradients.

Table~\ref{tab:main_qat} presents the performance of B158 models and their LGR-enhanced counterparts. For reference, the results of the FP16 models are also included. It is evident that the models trained with LGR exhibit better performance on the majority of benchmarks. Although lower scores on some tasks reduce the overall average, this does not negate the effectiveness on the majority of benchmarks. We also observe that the 0.4B model outperforms the 1.7B model on certain tasks. This is likely attributable to the sufficiency of training.

\section{Discussion}

\subsection{Quantized Weight Space is Anisotropic}

\begin{figure*}[t]
\centering
  \begin{minipage}{0.45\textwidth}
  \centering
  \includegraphics[width=0.85\linewidth]{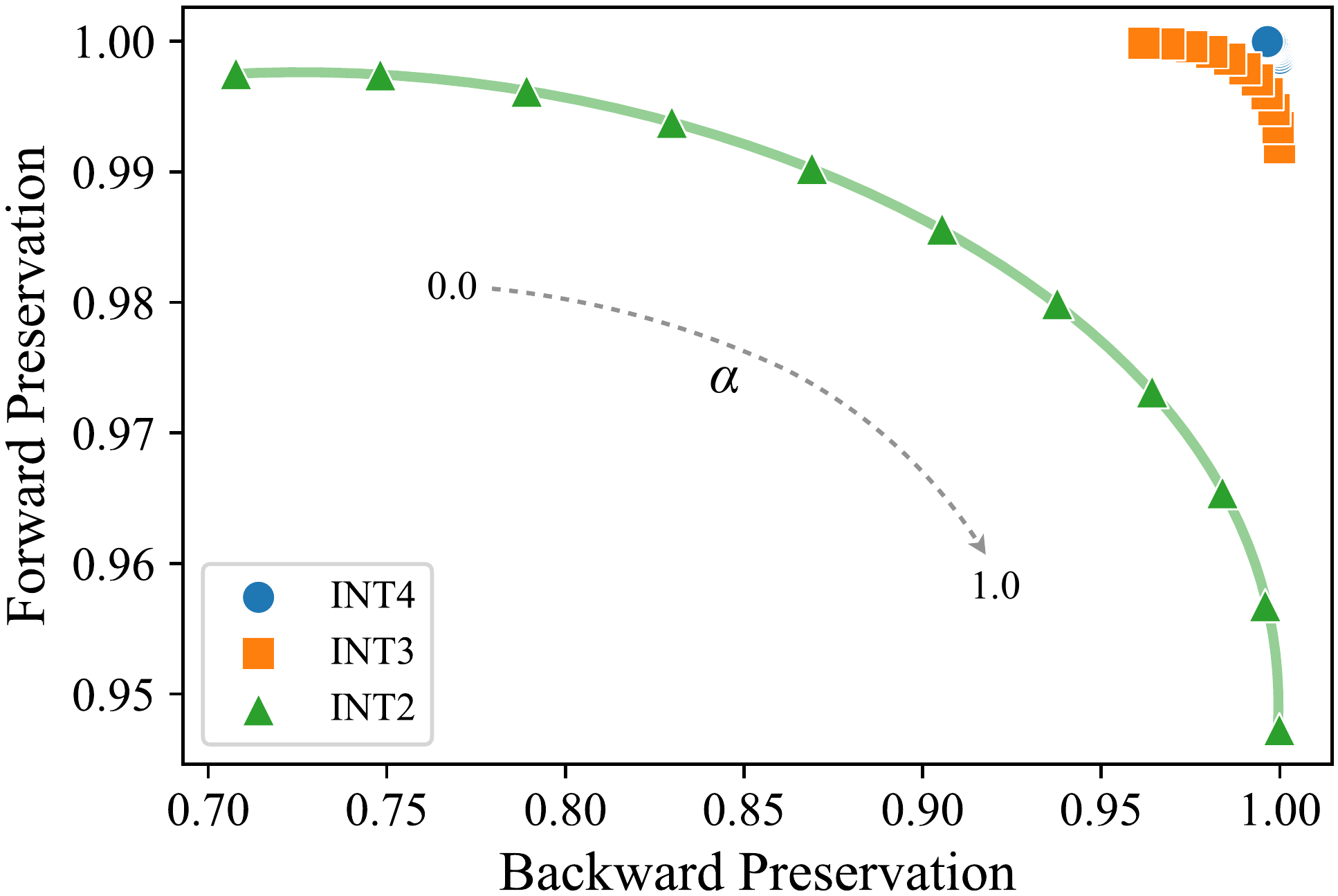}
  \caption{Forward / backward preservation.}
  \label{fig:f_b}
  \end{minipage}
  \hspace{10mm}
  \begin{minipage}{0.45\textwidth}
  \centering
  \includegraphics[width=0.8\linewidth]{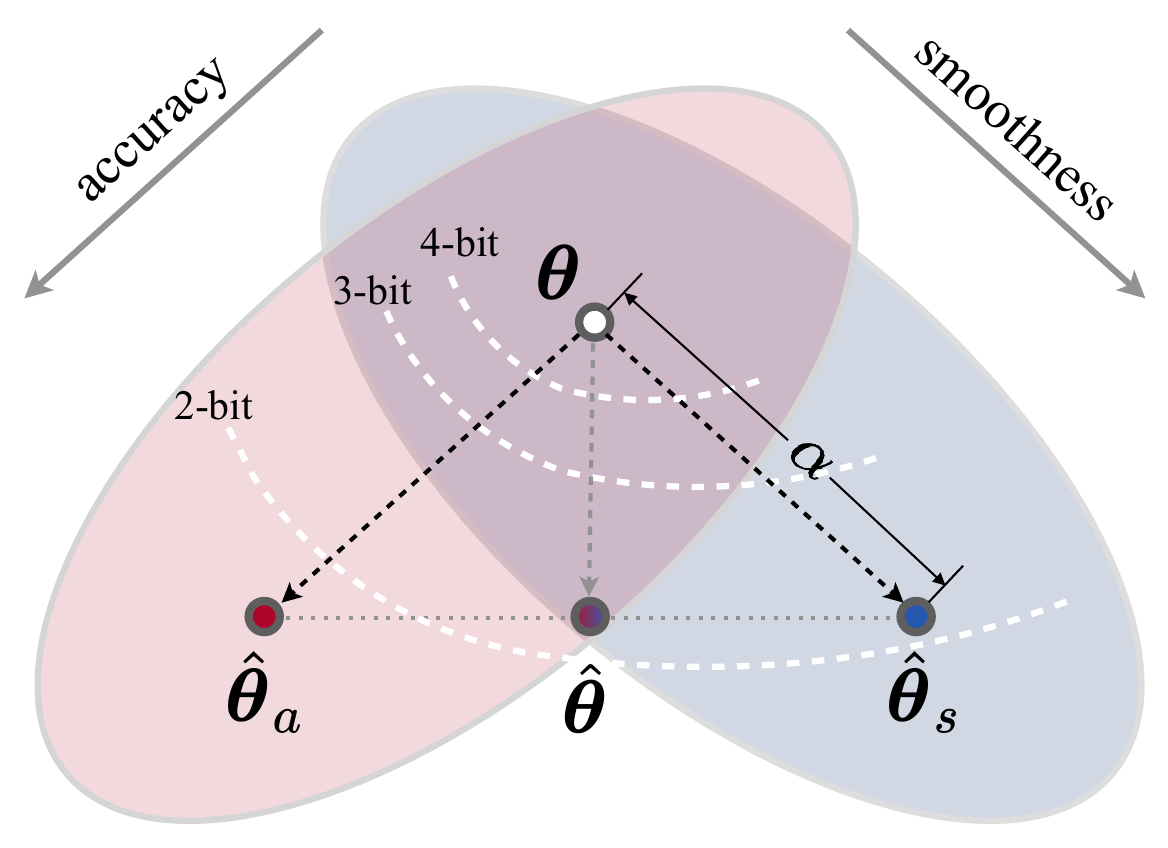}
  \caption{Anisotropy of the weight space.}
  \label{fig:illus}
  \end{minipage}
\end{figure*}

Although we are still unable to derive a high-quality closed-form solution to Equation~\ref{eq:core}, we can study the change in solution properties from two extremes: fitting accuracy and smoothness preservation. Specifically, we investigate the quantization of the \texttt{q\_proj} weights in the 0-th layer of LLaMA-2-7B and separately optimize the two components of the objective in Equation~\ref{eq:core}. We optimize $\|\mathbf{WX}-\hat{\mathbf{W}}\mathbf{X}\|_F^2$ to obtain $\hat{\mathbf{W}}_a$ using the GPTQ algorithm, and optimize $\|\mathbf{W}^\top\mathbf{G}-\hat{\mathbf{W}}^\top\mathbf{G}\|_F^2$ to obtain $\hat{\mathbf{W}}_s$. By constructing $\hat{\mathbf{W}}=(1-\alpha)\hat{\mathbf{W}}_a+\alpha\hat{\mathbf{W}}_s$ and varying the value of $\alpha$, we obtain a series of intermediate points between $\hat{\mathbf{W}}_a$ and $\hat{\mathbf{W}}_s$. Even if these intermediate points are not solutions to Equation~\ref{eq:core}, their trends clearly reveal how the properties of the solution space change along different directions. We use cosine similarity to measure how well the quantized outputs $\hat{\mathbf{W}}\mathbf{X}$ and $\hat{\mathbf{W}}^\top\mathbf{G}$ preserve the original outputs $\mathbf{WX}$ and $\mathbf{W}^\top\mathbf{G}$, as shown in Figure~\ref{fig:f_b}. At 4-bit quantization, the difference between solutions $\hat{\mathbf{W}}_a$ and $\hat{\mathbf{W}}_s$ is negligible, with no clear difference between fitting accuracy and smoothness preservation. However, as the quantization bit-width decreases to 3-bit and even 2-bit, the \textit{Pareto frontier} shifts rapidly, making it difficult to optimize fitting accuracy and smoothness simultaneously. In addition, backward smoothness is more sensitive to low-bit quantization than forward accuracy. At 2-bit, when only fitting accuracy is considered (i.e., $\hat{\mathbf{W}}=\hat{\mathbf{W}}_a$), the backward preservation score drops sharply to $0.70$. The shift of the Pareto frontier and the different sensitivities along the two directions suggest that an appropriate trade-off exists between the two extreme solutions.

Model quantization can be viewed as shifting the original weights along specific directions. The above analysis suggests that quantization performance depends on the choice of weight shift directions. As illustrated in Figure~\ref{fig:illus}, in high-precision regimes, both accuracy and smoothness can simultaneously achieve optimality. However, under large deviations between $\hat{\boldsymbol{\theta}}$ and $\boldsymbol{\theta}$ at lower bit-width, optimizing fitting accuracy alone leads to the loss of smoothness. The weight anisotropy in extreme quantization suggests that smoothness must be considered explicitly.




\subsection{Accuracy + Smoothness = $\varnothing$?}
\label{subsec:solution}

Let $\Delta\mathbf{W}=\mathbf{W}-\hat{\mathbf{W}}$, and $\|\mathbf{WX}-\hat{\mathbf{W}}\mathbf{X}\|_F^2$ can be simplified to $\min\|\Delta\mathbf{W}\mathbf{X}\|_F^2$. Given that $\mathbf{X}$ adheres to a certain distribution, a $\Delta\mathbf{W}$ exists that minimizes this term, which partially explains the viability of quantization. If we subsequently impose the stronger constraint of $\|\mathbf{W}^\top\mathbf{G}-\hat{\mathbf{W}}^\top\mathbf{G}\|_F^2$ (i.e., $\min\|\Delta\mathbf{W}^\top\mathbf{G}\|_F^2$), does a valid $\Delta\mathbf{W}$ still persist? Does this imply that the original FP16 model is the sole candidate capable of simultaneously achieving both accuracy and smoothness? The answer seems to be \textit{no}. This stems from the fact that the two optimization processes essentially involve finding the rows of $\Delta \mathbf W$ in the null spaces of $\mathbf X$ and the columns must lie in the null space of $\mathbf G$. A joint solution exists provided that $\mathcal N(\mathbf X^\top)\cap\mathcal N(\mathbf G^\top)\neq\{\mathbf 0\}$. Indeed, the extreme sparsity of LLMs ensures that $\mathbf{X}$ and $\mathbf{G}$ are low-rank, making the condition:
\begin{equation}
    \mathrm{rank}(\mathbf X) + \mathrm{rank}(\mathbf G) < \min(d_{\text{in}}, d_{\text{out}})
\end{equation}
generally valid, though the feasible solution space becomes much narrower. This observation highlights the challenges that dual constraints impose on developing algorithms for extreme quantization.

\subsection{How Does Smoothness Help Quantized Models?}

Fitting accuracy improves model capability by maximizing the predicted probabilities of high-quality tokens. However, for a fixed weight bit-width, the achievable fitting accuracy is inherently limited. In contrast, the smoothness objective does not directly increase the probabilities of high-quality tokens, but instead improves model performance by influencing token rankings. This phenomenon can be understood from the trend of the rPPL-$k$ curves. To slow down the growth of rPPL and make the curve smoother, the model must rank high-quality tokens higher in the prediction order, even if their probability gains are not substantial.

\begin{figure}[tbp]
  \centering
  \includegraphics[width=0.60\columnwidth]{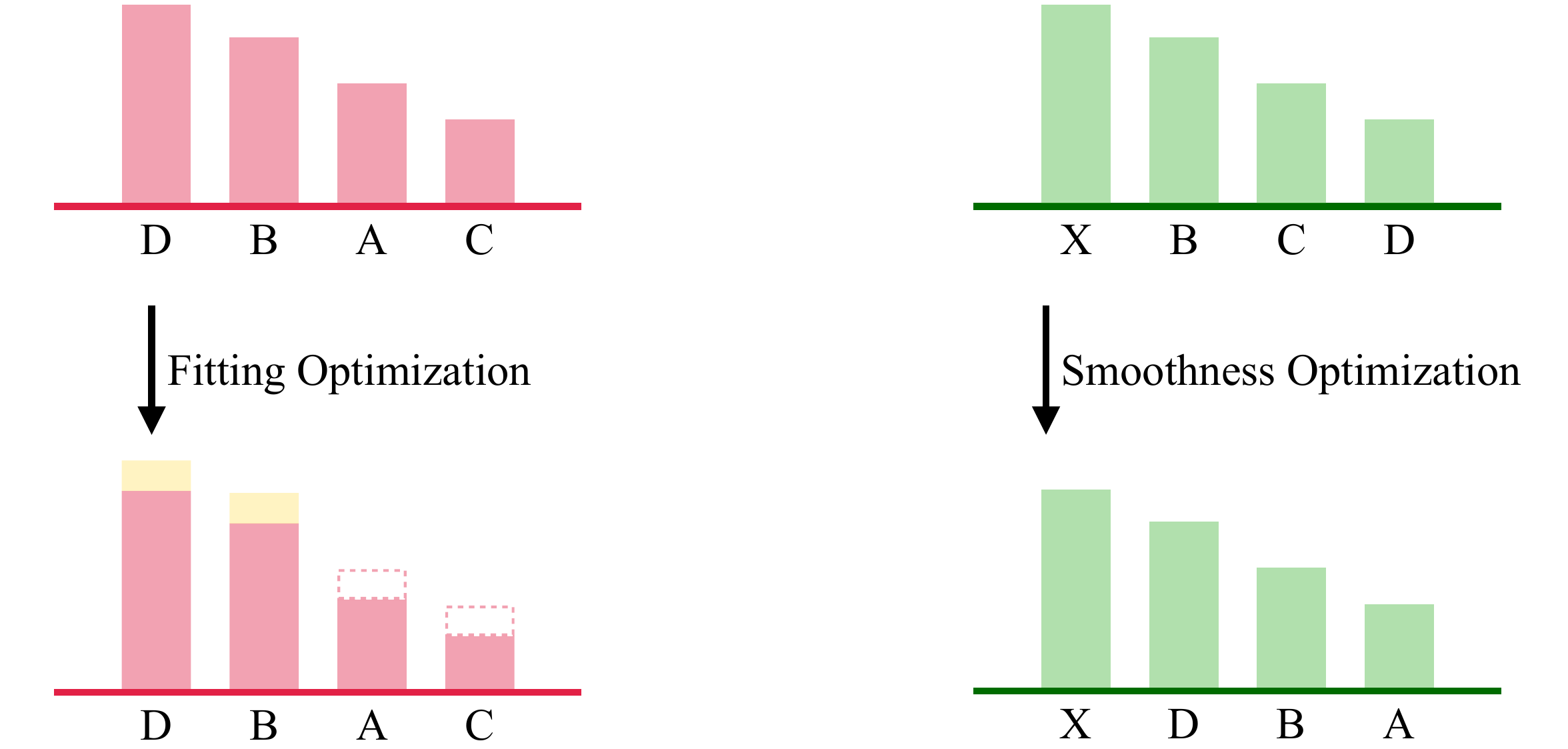}
  \caption{Comparison of fitting optimization and smoothness optimization. Given a prompt, the ideal ranking of predicted tokens is ``DBAC''.}
  \label{fig:diff}
\end{figure}

For example, the nine zero-shot tasks used in our evaluation are all multiple-choice benchmarks, where correctness is determined by the relative ranking among options A, B, C, and D. Suppose the correct answer is D. After introducing smoothness regularization, the model may change the prediction order from ``BCDA'' to ``DBAC'', as shown in Figure~\ref{fig:diff}. Although none of the four options ranks highly in the overall ranking, this change is still sufficient to turn an incorrect prediction into a correct one.

Therefore, what we would like to emphasize in this paper is that fitting accuracy and smoothness play different roles in model performance, and this distinction becomes particularly important under extreme quantization. However, one should never expect smoothness optimization alone to yield substantial performance gains. Fitting accuracy remains the primary factor determining quantization performance, while smoothness serves as a complementary enhancement.

\subsection{The Role of LayerNorm in Sub-2-bit Models}

Our training experience with sub-2-bit models suggests that introducing \texttt{LayerNorm} into linear layers is crucial, which is also emphasized in several previous studies~\citep{onebit2024,era2024}. The discussion in this paper helps explain the necessity of this practice. Beyond improving training stability, \texttt{LayerNorm} also smooths the backward gradients during training. Therefore, \texttt{LayerNorm} contributes indispensably to the performance of sub-2-bit models.

\section{Related Work}

\subsection{Model Quantization}

Model quantization compresses high-precision weights into low-bit representations. Post-training quantization (PTQ) converts a trained model to low-bit precision using optimization solvers~\citep{gptq2022,spqr2023}. Quantization-aware training (QAT) integrates quantization into training process to mitigate the adverse effects of reduced precision~\citep{llmqat2023,onebit2024}. Some studies also explore simultaneous weight and activation quantization~\citep{spinquant2025,smooth2023}. Currently, the widely used lossless quantization level is INT4. Quantization below this level is considered extreme, but it often leads to degraded model performance~\citep{quip2024,billm2024,era2024,arb2025,crvq2025}. Nearly all existing extreme quantization focuses primarily on improving representation accuracy to reduce forward computation loss.

\subsection{Network Smoothness}

Lipschitzness is the most direct smoothness metric in neural networks~\citep{spectrally2017}. However, despite its concise definition in machine learning, accurately estimating the Lipschitz constant for neural networks is highly challenging~\citep{lipschitz2018}. Numerous studies attempt to approximate network Lipschitz constants, yet they typically suffer from two limitations. First, they fail to effectively address the exponential growth of estimated bounds with network depth~\citep{lipschitz2018,quantitative2022}. Second, most prior work relies on early architectures~\citep{regularisation2021,skew2021,lipschitz2021,lipschitz2019,some2024}, such as ResNet. In contrast, the Lipschitzness of transformer~\citep{lipschitz2021self,lipsformer2023} remains relatively underexplored. Although early studies in vision~\citep{regularisation2021} and robustness~\citep{training2021} investigate the impact of smoothness on model performance via proxies, conclusions on smoothness in LLMs remain scarce.

\section{Conclusion}

In this work, we highlight smoothness as a important but overlooked objective in extreme quantization. Building on input-gradient analysis and sequence neighborhood modeling, we introduce LGP for PTQ and LGR for QAT as simple smoothness-preserving instantiations, and advocate explicitly incorporating smoothness into future quantization design.

\bibliographystyle{abbrvnat}
\bibliography{neurips_2026}






\appendix

\vspace{10mm}
\section{Appendix}

\subsection{Smoothness in Training Process}
\label{app:smoothness}

\begin{figure}[ht]
  \centering
  \begin{subfigure}{\columnwidth}
    \centering
    \includegraphics[width=0.5\linewidth]{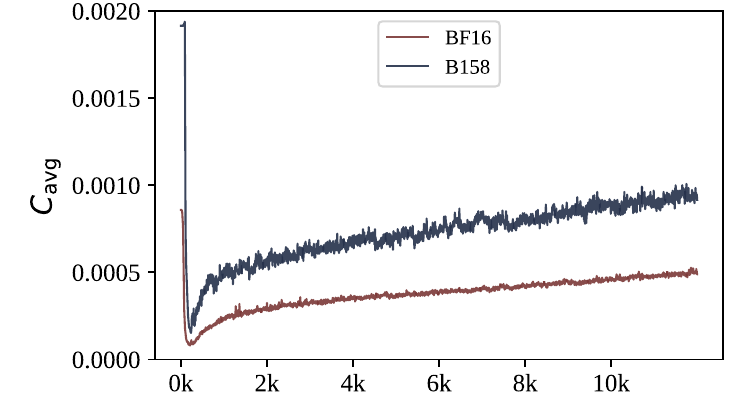}
    \caption{BF16 model and B158 model}
    \label{fig:smooth_bf16}
    \vspace{2mm}
  \end{subfigure}
  \begin{subfigure}{\columnwidth}
    \centering
    \includegraphics[width=0.5\linewidth]{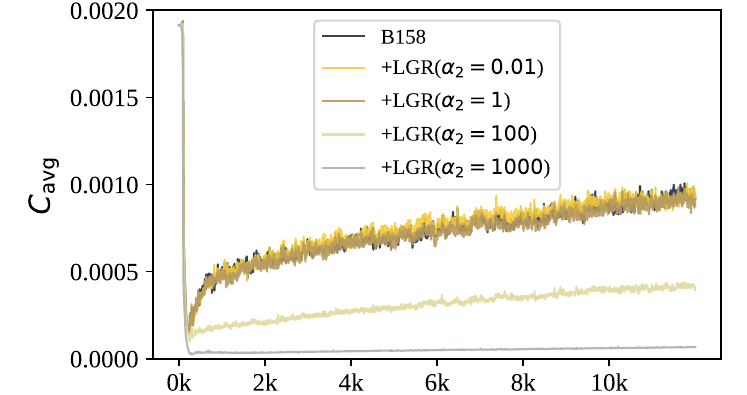}
    \caption{B158 model and different LGR settings}
    \label{fig:smooth_b158}
  \end{subfigure}
  \caption{Evolution of $C_\text{avg}$ during training across different models and settings. We use 0.4B model in this experiment.}
  \label{fig:smooth}
\end{figure}

In this section, we discuss several topics related to smoothness-aware training, reflecting on insights and conjectures derived from our experiments. We primarily address the following three questions:

\noindent\textbf{\textit{a) Why is RMSNorm excluded from the BitNet b1.58 model in our QAT experiments?}}

In the standard BitNet b1.58 architecture~\citep{era2024}, the activations entering the linear layer are first normalized via \textit{RMSNorm} before performing matrix multiplication with the ternary weights. This process is defined as follows:
\begin{equation}
\begin{aligned}
\mathbf{X}^{'}&=\mathrm{RMSNorm}(\mathbf{X})\\
\mathbf{W}^{'}&=\mathrm{Ternarize}(\mathbf{W})\\
\mathbf{Y}&=\mathbf{W}^{'}\mathbf{X}^{'}.\notag
\end{aligned}
\end{equation}
In our LGR implementation, we deliberately excluded \textit{RMSNorm} as a controlled variable strategy. It is well established that BitNet models incorporating \textit{RMSNorm} achieve performance comparable to full-precision baselines. This effectiveness stems from two factors: first, \textit{RMSNorm} normalizes activations during the forward pass, enhancing computation accuracy; second, it significantly flattens backward gradients, substantially improving ternary model smoothness.

However, this dual impact makes it difficult to isolate the specific benefits of smoothness. By removing \textit{RMSNorm}, we align the forward computation process with that of the full-precision model. Consequently, any observed gains can be attributed solely to smoothness. Furthermore, the success of \textit{RMSNorm} itself implicitly validates the accuracy-smoothness trade-off idea proposed in this paper.

\noindent\textbf{\textit{b) Is smoother training always better?}}

In QAT, maximizing smoothness is not unconditionally beneficial. Moreover, it is unrealistic to expect a ternary model to achieve parity with a full-precision model in both smoothness and performance simultaneously. The core idea of this work is to identify the smoothest candidate among the manifold of quantized solutions that have equivalent fitting accuracy.

The guiding principle is \textbf{moderate smoothing}: enhancing smoothness without compromising accuracy. We selected the hyperparameter $\alpha_2=0.01$. As shown in Figure~\ref{fig:smooth}, increasing $\alpha_2$ significantly improves model smoothness; at $\alpha_2=100$, the ternary model exhibits smoothness comparable to the full-precision baseline. However, this comes at a cost: due to over-smoothing, the accuracy degrades, causing perplexity to deteriorate from 97.4 to 130.6, similar to randomly regularization. Consequently, excessive smoothing must be avoided.

\noindent\textbf{\textit{c) How can smoothness be incorporated into industrial training practices?}}

Since the norm of input gradients is negligible relative to the training loss, introducing smoothness constraints in the early stages of training proves ineffective. Furthermore, adhering to the principle of moderate smoothing, LGR should not drastically alter the intrinsic smoothness of quantized model. Consequently, we recommend incorporating smoothness training only during the mid-to-late stages of the training process. This strategy may preserves pre-trained knowledge while conserving computational resources.

\begin{figure}[t]
  \centering
  \begin{subfigure}{\columnwidth}
    \centering
    \includegraphics[width=0.5\linewidth]{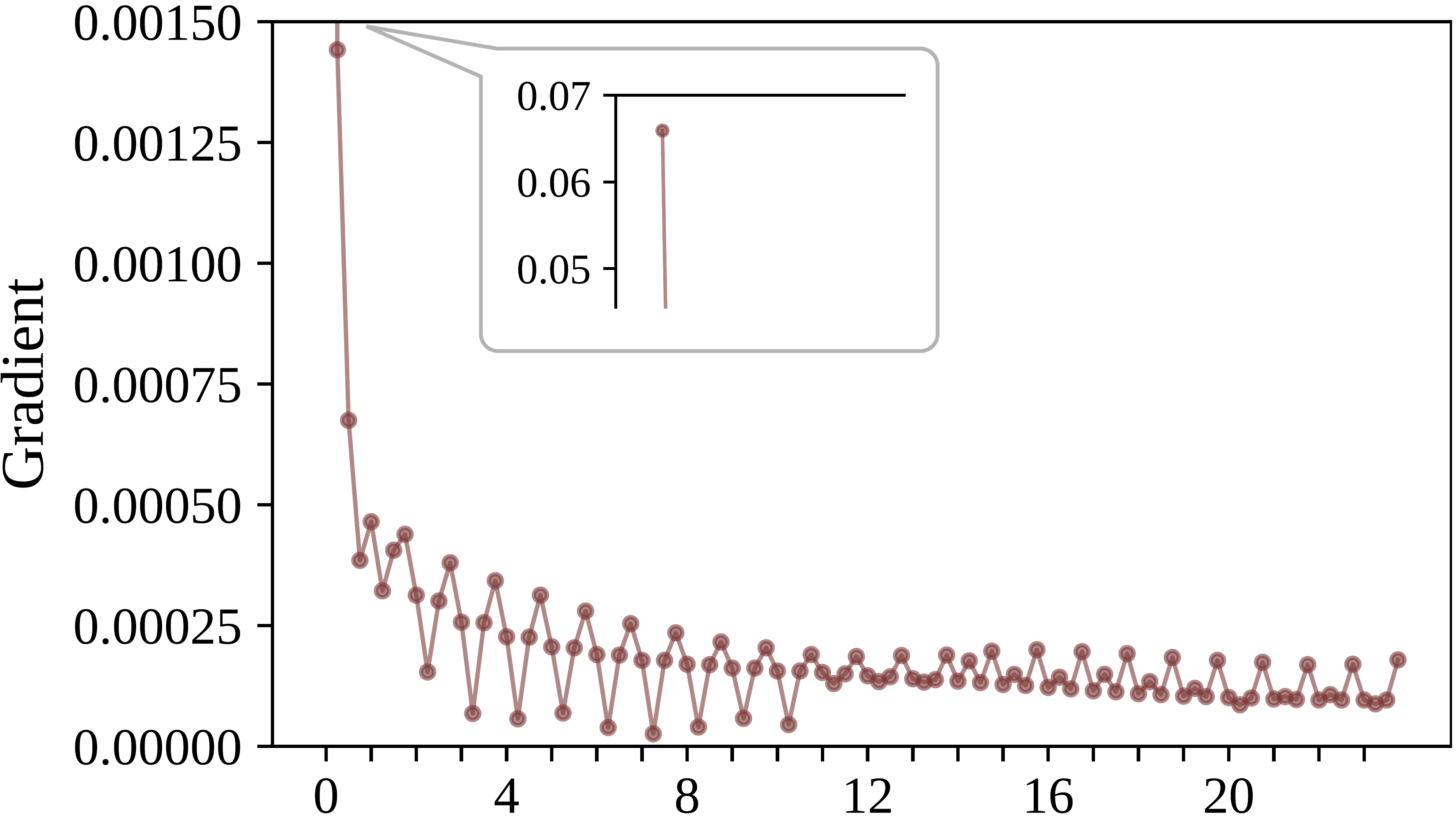}
    \caption{BitNet b1.58 model}
    \label{fig:grs_bitnet}
    \vspace{2mm}
  \end{subfigure}
  \begin{subfigure}{\columnwidth}
    \centering
    \includegraphics[width=0.5\linewidth]{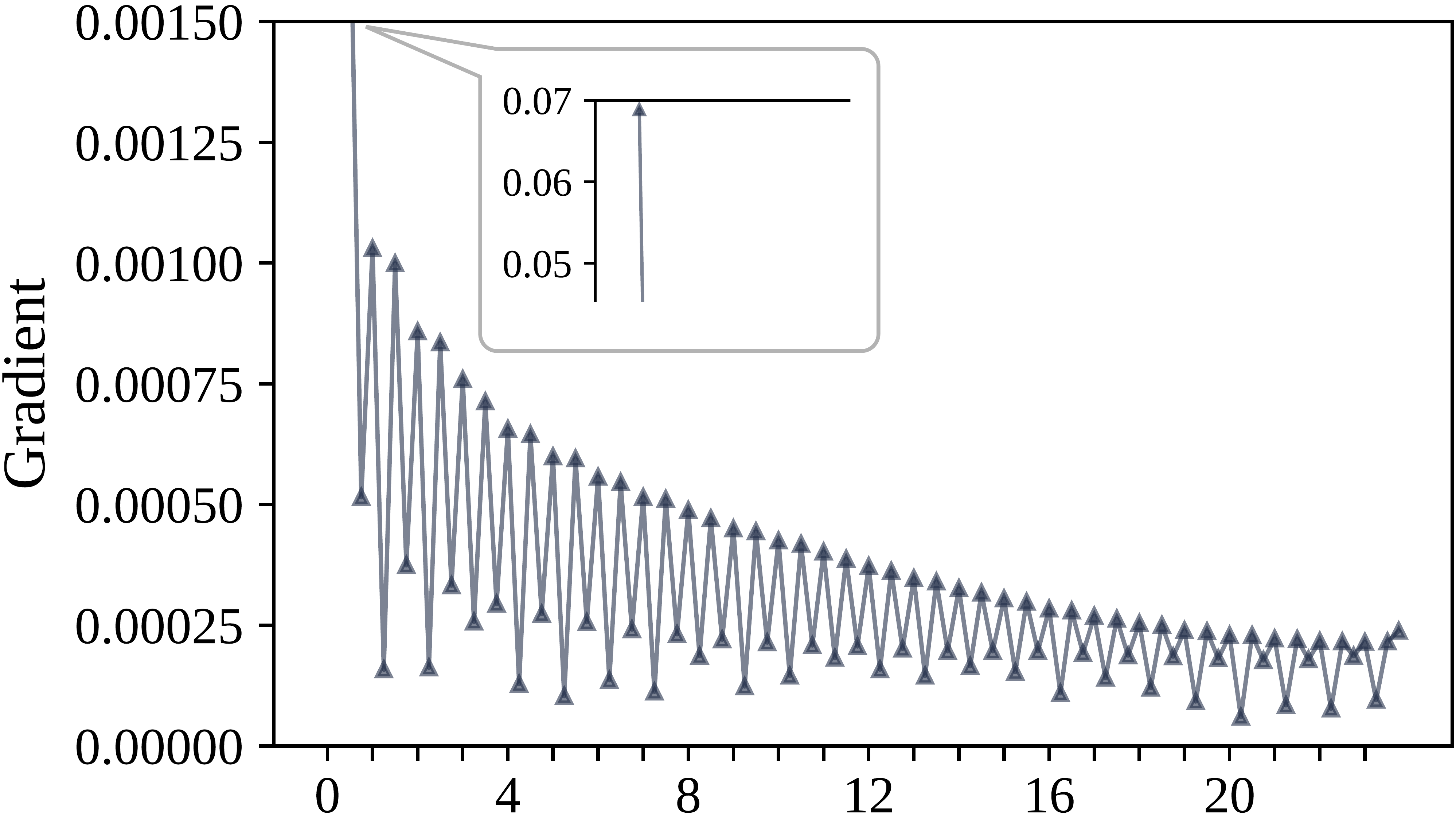}
    \caption{B158 model with frozen embedding}
    \label{fig:grs_freeze}
  \end{subfigure}
  \caption{Evolution of $C_\text{avg}$ during training when integrate \textit{RMSNorm} or freeze word embedding.}
  \label{fig:grs}
\end{figure}

\subsection{Gradient Ridge}
\label{app:gr}

We observe that the ``Gradient Ridge'' is not sporadic. It is independent of subtle architectural modifications or quantization settings. In Section~\ref{subsec:lgr}, we initially demonstrate its presence in both FP16 and B158 models. To reinforce the universality of this observation, we conducted similar tests on the BitNet b1.58 architecture with \textit{RMSNorm}.

As illustrated in Figure~\ref{fig:grs_bitnet}, even though the \textit{RMSNorm}-equipped model exhibits smoothness comparable to the FP16 baseline, the input gradient at 0-th layer still spikes to a very high magnitude. Note that all results reported thus far, including those in Section~\ref{subsec:lgr}, is derived under full-weight training settings. Furthermore, we test with using frozen FP16 embeddings within the B158 model. As depicted in Figure~\ref{fig:grs_freeze}, the conclusion holds firm.

Based on this evidence, we conclude that the \textit{Gradient Ridge} is by no means coincidental. Although its root cause remains elusive, we believe that exploring it could offer significant benefits for the interpretability of LLMs.

\subsection{Residual Quantization and Smoothness}
\label{app:residual}

Section~\ref{subsec:lgp} points out the challenge of optimizing the complete objective (Equation~\ref{eq:core}), as the two sub-objectives are orthogonal. Yet, Section~\ref{subsec:solution} clarifies that this goal is not impossible. It merely constrains the solution space.

\begin{figure}[t]
  \centering
  \includegraphics[width=0.35\columnwidth]{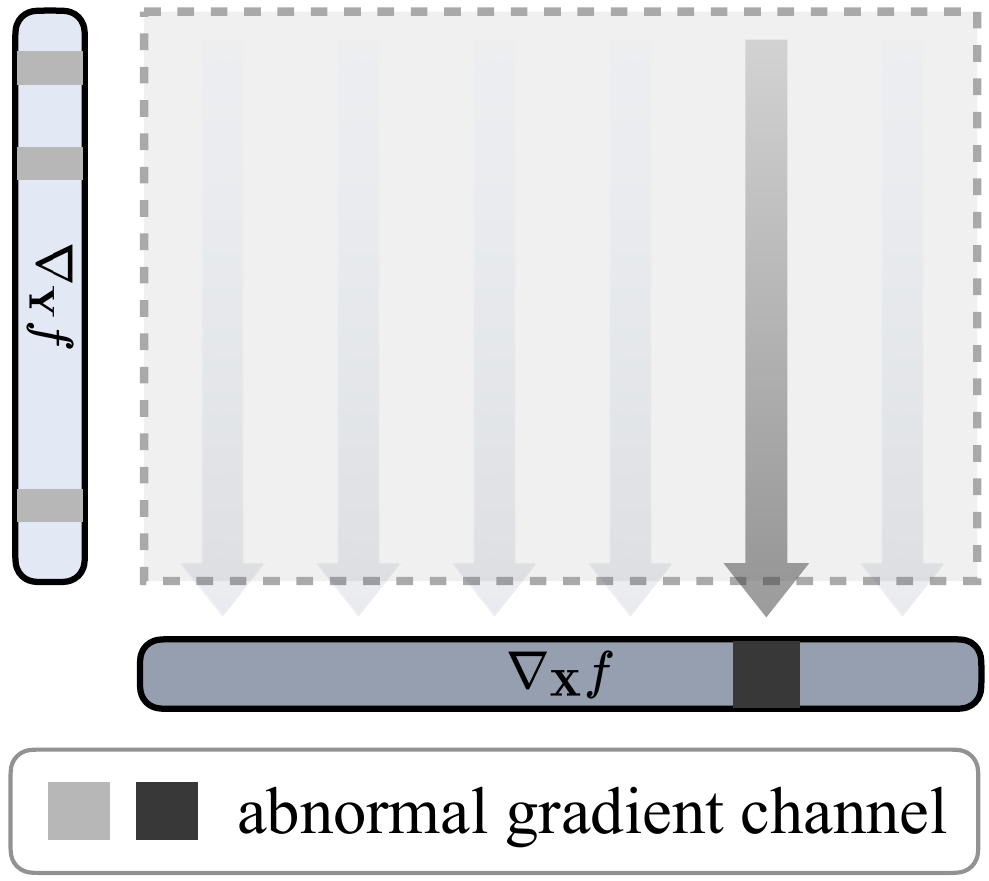}
  \caption{Important/abnormal gradient channels and relaxed optimization.}
  \label{fig:residual}
\end{figure}

While deriving a closed-form solution for Equation~\ref{eq:core} is challenging, it remains feasible to partially fulfill the smoothness objective while prioritizing accuracy. The extreme sparsity of LLM gradients underpins this feasibility. As illustrated in the Figure~\ref{fig:residual}, input and output gradients across linear layers remain negligible in the vast majority of channels. This implies that by prioritizing the precision of a select few columns, we can ensure that gradient propagation remains unimpaired. As a result, selecting key column vectors and applying high precision presents an intuitively viable strategy. Here, the objective in $\min_{\hat{\mathbf{W}}} \|\mathbf{W}^\top \mathbf{G} - \hat{\mathbf{W}}^\top \mathbf{G} \|^2_F$ is relaxed, effectively giving an approximate solution.

In fact, similar ideas have already emerged in the field of extreme quantization~\citep{billm2024,crvq2025,arb2025}. Methods like \textit{residual quantization} select a subset of pivotal column vectors to fit their quantization errors, significantly enhancing extreme compression performance. The distinction between our proposal and such methods lies in the selection criteria. Residual quantization evaluates column importance based on activation error, defined as $I_i = (w_i-\hat{w}_i)x_i$. In contrast, our relaxed formulation utilizes the magnitude of backward gradients $\nabla_{x_{i}} f$ as the importance metric. Combining these two indicators may pave the way for designing even more potent extreme compression algorithms, which is a direction we leave for future research.

\subsection{Experimental Settings}
\label{subsec:setup}

\subsubsection{LGP}
\label{subsec:setup1}

We evaluate the effectiveness of introducing LGP on two mainstream models, LLaMA-2 and Qwen-3. The model sizes range from 0.6B to 13B.

\paragraph{Baselines} We select the original model, 2-bit GPTQ, and 2-bit OmniQuant as our baselines. The original model serves as a reference to quantify the performance gap induced by quantization. By comparing OmniQuant with GPTQ, we highlight the accuracy gains achieved by the learnable weight clipping. Our method is designed to demonstrate the effectiveness of introducing LGP for joint optimization. For calibration, all methods utilize 128 sequences from the C4 dataset, each with a length of 2048. The quantization group size is set to 128.

\paragraph{LGP} For all models, regardless of whether LGP is incorporated, the layer-wise distillation process is conducted for 40 epochs. The coefficient $\alpha_1$ is typically set to be $1e^K$ where $K$ is a positive integer, as listed in Table~\ref{tab:main}.

\paragraph{Evaluation} To assess the performance of the baselines and LGP, we calculate perplexity using sequences randomly sampled from Wikitext2~\citep{wiki22016} and C4~\citep{C42020}. Additionally, we report zero-shot accuracy across a range of tasks, including Winogrande~\citep{winogrande2021}, Hellaswag~\citep{hellaswag2019}, PIQA~\citep{piqa2020}, BoolQ~\citep{boolq2019}, ARC~\citep{arc2018}, OBQA~\citep{obqa2018}, MathQA~\citep{mathqa2019}, and RTE~\citep{glue2018}.

\subsubsection{LGR}
\label{subsec:setup2}

We evaluate the FP16 baseline against the standard B158 model and its smoothed counterpart. We train models of three different sizes on the OpenWebText2 dataset, and their configurations are listed in Table~\ref{tab:models}, with detailed training settings provided in the Appendix. We fix $\alpha_2 = 0.01$ for all runs. The evaluation protocol aligns with that of Section~\ref{subsec:setup1}.

\begin{table*}[t]
\centering
\renewcommand{\arraystretch}{0.9}
\begin{tabular}{c|ccccccc}
\toprule
size & $n_\text{layer}$ & $n_\text{head}$ & $d_\text{hidden}$ & $d_\text{inter}$ & steps & $lr$ \\ 
\midrule
0.4B & 16 & 16 & 1024 & 4096 & 300k & 2e-4 \\
1.0B & 24 & 24 & 1536 & 6144 & 400k & 2e-4 \\
1.7B & 24 & 32 & 2048 & 8192 & 400k & 2e-4 \\
\bottomrule
\end{tabular}
\caption{Configurations of different model sizes.}
\label{tab:models}
\end{table*}

\subsection{Ablation}
\label{subsec:ablation}

\begin{figure}[t]
  \centering
  \includegraphics[width=0.50\columnwidth]{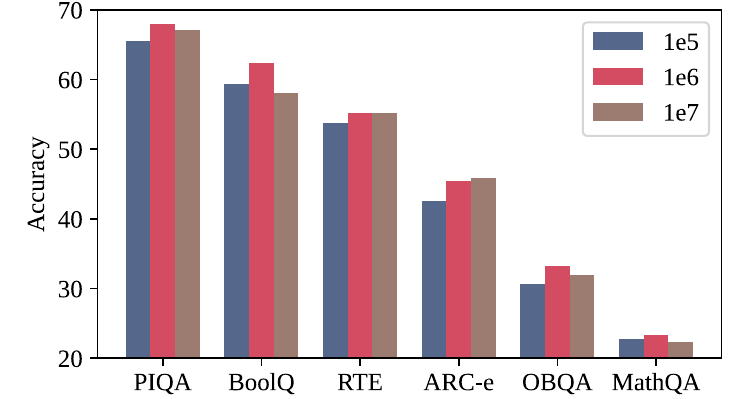}
  \caption{Comparison of zero-shot accuracy under different values of the coefficient $\alpha_1$.}
  \label{fig:coef}
\end{figure}

LGP incorporates a parameter $\alpha_1$ to balance the objectives of fitting fidelity and smoothness. An insufficiently small $\alpha_1$ renders LGP ineffective, while an overly large value disrupts the fitting process, leading to accuracy degradation. To avoid these issues, we employ a heuristic method for selecting $\alpha_1$. Given that the norm of the backward gradients is smaller than the fitting loss by multiple orders of magnitude, we choose a sufficiently large $\alpha_1$ to bring both terms to a comparable scale. For example, for LLaMA-2-7B, we perform a search across magnitudes ranging from $1e4$ to $1e8$. The difference of varying $\alpha_1$ is shown in Figure~\ref{fig:coef}.

\begin{table*}[t]
\centering
\begin{tabular}{c|ccccccc}
\toprule
LGR & C4 & Wino & PIQA & ARC-e & ARC-c \\ 
\midrule
$\nabla_{\mathbf{x}^{(1)}} f$ & 68.11 & 50.83 & 54.13 & 29.42 & 23.29 \\
$\nabla_{\mathbf{x}^{(0)}} f$ & 70.29 & 48.22 & 53.97 & 29.00 & 22.78 \\
\bottomrule
\end{tabular}
\caption{Ablation of different input gradients. We report the 0.4B model training with 400k steps, and $\alpha_2 = 0.1$.}
\label{tab:ablation2}
\end{table*}

Table~\ref{tab:ablation2} shows the superiority of utilizing input gradients from the 1-st layer. In contrast, relying on the 0-th layer may have a detrimental impact on training results. We attribute this potential failure to the corruption of embedding representations.

\subsection{Details on Baselines}
\label{app:baselines}

In this section, we provide supplementary implementation details for the evaluated methods. All code will be released upon de-anonymization of the paper.

\paragraph{GPTQ} We adhere to the default configurations provided in the official open-source repositories. Specifically, we set the random seed for data sampling to 0 and employ a Hessian damping factor of 0.01. Moreover, we utilize asymmetric quantization, activation-aware channel reordering, and true sequential quantization.

\paragraph{OmniQuant} We set the random seed for data sampling to 0. The learning rate for LWC is configured at 0.01, and the model is trained for alignment for 40 epochs using AdamW optimizer. Additionally, we employ asymmetric quantization and an auxiliary loss in their official code.

\paragraph{LGP} Following OmniQuant, we quantize layers sequentially in a shallow-to-deep manner. For each layer, we first perform a full forward and backward pass through the entire model to compute the input and output gradients of this layer. Subsequently, we calculate the smoothness loss of the quantized model and add it to the original OmniQuant loss. Other configurations remain identical to those in OmniQuant.

\paragraph{LGR} We conduct the training on 8 NVIDIA A800 GPUs. The setup utilizes a per-device batch size of 1 with 16 gradient accumulation steps. We employ the AdamW optimizer with hyperparameters $\beta_1=0.9, \beta_2=0.95$, and a weight decay of 0.1. The learning rate follows a cosine schedule with a 1000-step warmup. To help for reproducibility, random seeds for Python, NumPy, PyTorch, and CUDA are all fixed at 1234.

\subsection{Limitations}

Although we propose a novel smoothness-aware perspective to complete the optimization objectives for extreme compression, we acknowledge three main limitations that point toward future directions.

Firstly, our proposed methods (LGP and LGR) function as foundational baselines. We prioritized simplicity and versatility to demonstrate the validity of the smoothness hypothesis. Consequently, these methods may not represent the optimal mathematical solution for the joint optimization problem we formulated. We anticipate that future work can build upon our dual-constraint analysis to design more advanced solvers.

Secondly, while we provide intuitive explanations like the ``Gradient Ridge'', its underlying causes are not yet fully understood. We regard this work as a stepping stone that invites the community to delve deeper into its underpinnings, which may contribute to the explainability of LLMs.

Lastly, constrained by available computational resources, our experiments are confined to models of moderate scale and trained with a limited number of steps on a restricted dataset. While we observe consistent improvements within this scope, the behavior of smoothness constraints under the regime of massive-scale pre-training remains an open question. We hope that future work can further scale these experiments.



\end{document}